\pdfoutput=1
\documentclass[10pt,twocolumn,letterpaper]{article}
\usepackage[T1]{fontenc}
\usepackage[utf8]{inputenc}
\usepackage{lmodern}
\usepackage{microtype}
\usepackage[
  letterpaper,
  top=0.75in,
  bottom=0.80in,
  left=0.70in,
  right=0.70in,
  columnsep=0.24in
]{geometry}
\usepackage{graphicx}
\usepackage{booktabs}
\usepackage{multirow}
\usepackage{array}
\usepackage{amsmath}
\usepackage{amssymb}
\usepackage{bm}
\usepackage{mathtools}
\usepackage[authoryear,round]{natbib}
\usepackage[font=small,labelfont=bf]{caption}
\usepackage{xurl}
\usepackage[hidelinks]{hyperref}
\newcommand{\bestscore}[1]{\textbf{#1}}
\newcommand{\secondscore}[1]{\underline{#1}}

\title{Verifiable Memory: Learning Unified Memory Management
with Local and Global Verifiers for Large Language Model Agents}
\author{%
Xiaolong Sun\(^{1}\)\quad
Qichao Wang\(^{2}\)\quad
Hangyu Li\(^{3}\)\quad
Liang Chen\(^{1,*}\)\\[3pt]
\(^{1}\)Sun Yat-Sen University, Guangzhou, China\\
\(^{2}\)Nanyang Technological University, Singapore\\
\(^{3}\)Tencent, Shenzhen, China\\[3pt]
\texttt{sunxlong@mail2.sysu.edu.cn}\quad
\texttt{qichao001@e.ntu.edu.sg}\\
\texttt{masonhyli@tencent.com}\quad
\texttt{chenliang6@mail.sysu.edu.cn}\\
\(^{*}\)Corresponding author
}
\date{}

\begin{document}
\maketitle

\begin{abstract}
Large language model (LLM) agents must retain reusable information,
control a bounded active context, and recover earlier evidence during
long-horizon interaction. Existing methods commonly optimize long-term
memory (LTM) and short-term memory (STM) separately, while unified
policies are often trained primarily with trajectory-level feedback,
which provides weak credit for individual memory decisions. We present
Verifiable Memory (VerMem), a framework that represents LTM, active
context, and episodic history as distinct states and controls them with
one memory operation policy. Seven atomic operations let
the policy add, revise, or soft-delete LTM entries; retrieve LTM into the
active context; filter or summarize the active context; and restore
selected episodic fragments. VerMem is initialized by supervised
fine-tuning and trained with a three-stage reinforcement-learning
curriculum. The local verifier scores executable memory transitions, and
a global verifier assesses evidence coherence and terminal-memory
consistency after task completion. These scores are combined with
programmatically computed task, evidence-recall, efficiency, and
constraint signals through hierarchical credit assignment. The
verifiers are used only during training. Across five benchmarks and two
LLM backbones, VerMem achieves the best result on the vast majority of
reported metrics and consistently outperforms strong memory baselines.
Under controlled online-token budgets on three interactive benchmarks,
it also achieves the strongest efficiency--performance frontier among
the compared methods. Code is available at ~\url{https://github.com/Sun-SYSU-24/VerMem}.
\end{abstract}


\section{Introduction}

In long-horizon tasks involving multi-step interaction, tool use,
and complex reasoning, the performance of large language model
(LLM) agents depends not only on single-step generation, but also
on retaining and using task-relevant information across decisions.
Prior work uses \textit{agent memory} to describe the information
that an agent can attend to and use at a given time
\citep{yu2026agemem}. Such information includes stored past
executions and state representations formed from interaction
history \citep{xiong2025memorymanagement,goodyear2025state}.
Agent memory is commonly divided into long-term memory and
short-term memory \citep{yu2026agemem}. Long-term memory (LTM)
persistently stores user- or task-specific knowledge for reuse in
later turns, stages, or tasks
\citep{jiang2024long,zhong2024memorybank}. Short-term memory
(STM) contains information in the current input context and
supports ongoing reasoning, action selection, and context control
\citep{wu2025human,gao2025efficient}. LTM preserves information
across time, while STM keeps the current reasoning state usable.
Their coordination is therefore central to long-horizon agent
reasoning.

Existing systems often optimize only one side of this
process. STM-oriented methods organize or compress the
current reasoning state \citep{qian2026memobrain,li2025cam},
whereas LTM-oriented methods build external stores for
persistent recall \citep{zhong2024memorybank,xu2025amem}.
Stored information, however, affects a decision only after
it enters the active context at the right time, and reusing
task-misaligned experience can propagate
errors \citep{xiong2025memorymanagement}.
The AgeMem policy introduced by \citet{yu2026agemem} places LTM and STM
operations in one policy, but its STM actions focus on
retrieval, filtering, and summarization. Long-horizon tasks
may also require an earlier observation, tool output, or
intermediate conclusion to return from episodic history after
it has left the active context. This operation differs from
retrieving persistent knowledge and from compressing recent
context.

Learning such behavior introduces a second difficulty.
\citet{yan2025memoryr1}, \citet{zhou2025mem1}, and
\citet{yu2026agemem} optimize memory decisions
with downstream outcomes or trajectory-level feedback.
These signals compare complete rollouts, but do not identify
which atomic transition was useful or harmful. Process
supervision and transition-level credit improve intermediate
feedback \citep{setlur2024rewardingprogress,
luo2025agentlightning}, while
\citet{team2026mirothinker} combines local and global verification
during inference. Operation-level and trajectory-level verification
have not been jointly used to train a unified memory operation policy.

These limitations leave three challenges for a unified and
trainable memory operation policy. \textbf{(A) Heterogeneous memory
coordination.} LTM determines what should be stored, revised, or
removed. STM determines what should enter, remain in, or leave the
active context. The two memory types operate over different states
and time scales. Coordinating them within a shared decision process
while preserving their functional boundaries remains difficult.
\textbf{(B) Historical-context recovery.} The current prompt is
only a budgeted view of the full task history. Early observations,
tool outputs, and intermediate conclusions may become relevant
again after the subgoal changes. Recovering the relevant fragments
without reintroducing large amounts of irrelevant history is
therefore nontrivial. \textbf{(C) Multi-granularity credit
assignment.} The utility of a memory operation is often delayed
and depends on later operations. The correct write may become useful
only after a later retrieval, while an incorrect filter may cause
failure several steps later. Training must assign
operation-specific credit while preserving alignment with the
complete task objective.

To address these challenges, we propose \textit{Verifiable Memory
(VerMem)}, shown in Figure~\ref{fig:VerMem_overview}. VerMem preserves LTM,
active context, and episodic history as distinct states, while
one memory operation policy coordinates LTM maintenance,
STM control, and historical-context recovery through seven
atomic tools. Training uses a supervised fine-tuning (SFT) warmup followed by a
three-stage reinforcement learning curriculum for LTM,
STM, and their joint use. The local verifier evaluates each
realized memory transition, and the global verifier evaluates
the completed trajectory. Their separately normalized
advantages are combined at every memory decision. Both verifiers are
removed during inference. Across five long-horizon benchmarks and two
backbones, VerMem consistently improves task performance and achieves a
stronger efficiency--performance frontier.

The main contributions of this work are summarized as follows:
\begin{itemize}
    \item \textbf{We propose \textit{VerMem}, a unified agent
    memory management framework.}
    VerMem coordinates LTM maintenance and STM control
    through a single memory operation policy while preserving
    distinct states for LTM, active context, and episodic history.
    Its atomic action space further supports episode-level context
    selection.

    \item \textbf{We introduce local--global verifier-guided
    hierarchical credit assignment.}
    \textit{VerMem} constructs operation-level local advantages
    and trajectory-level global advantages from stateful
    multi-step rollouts. The two signals are combined at each
    memory decision, allowing operation quality and final task
    utility to jointly guide policy updates.

    \item \textbf{We conduct systematic evaluations across
    diverse complex tasks.}
    We compare \textit{VerMem} with strong memory baselines on five
    benchmarks and evaluate its efficiency--performance trade-off
    under controlled online-token budgets. Ablations further examine
    unified LTM/STM management, local and global credit signals, and
    the multi-component reward design.
\end{itemize}

\section{Related Work}

\subsection{Long-Term Memory (LTM)}

Long-term memory research studies how agents preserve, organize,
and reuse historical information beyond the active context.
\citet{langchain2025langmem} describes LangMem as a set of modular primitives
for recording, searching, and consolidating persistent knowledge.
\citet{zhong2024memorybank} introduce MemoryBank for long-term recall
and user profiling through continual memory updates.
The Mem0 framework of \citet{chhikara2025mem0} extracts, consolidates,
and retrieves
salient information from extended conversations, while
the A-Mem framework of \citet{xu2025amem} organizes memory through dynamic
indexing, linking, and evolution. These systems improve
persistent-state construction, but their primary concern is how
information is stored and organized. Stored knowledge affects
current decisions only after it enters the active context at the
appropriate stage. Incorrect or task-misaligned experiences may
also propagate errors when reused \citep{xiong2025memorymanagement}.
This motivates joint control of persistent-memory maintenance and
active-context use rather than isolated optimization of LTM.

\subsection{Short-Term Memory (STM)}

Short-term memory research focuses on maintaining the active
context required by ongoing reasoning under a bounded context
budget. \citet{qian2026memobrain} organize reasoning
traces and transient tool outputs into a compact executive memory.
\citet{li2025cam} develop structured schemata and activate
query-relevant information for long-document understanding.
\citet{wu2025resum} periodically compress growing
interaction histories into concise reasoning states and further adapt agents to reason over these summaries with ReSum-GRPO.
These methods improve context usability, but they remain centered
on the current input, document, or ongoing trajectory. In
long-horizon tasks, useful evidence may remain in the episodic
history after it leaves the active context. Retrieval, filtering,
and summarization alone do not fully specify which earlier episode
should return to the current reasoning state. \textit{VerMem}
addresses this gap by selecting task-relevant fragments from
episodic history and materializing them into the active context as
an explicit STM decision.

\subsection{Learning Memory Policies with Verification}

Reinforcement learning turns memory management into a
learnable decision process. \citet{yan2025memoryr1} learn structured
updates, \citet{huo2026atommem} expose atomic memory operations,
\citet{zhou2025mem1} jointly learn consolidation and reasoning, and
\citet{yu2026agemem} train a unified LTM/STM policy. These methods
primarily optimize memory behavior with
answer-level or trajectory-level objectives. Process supervision
evaluates intermediate reasoning
progress \citep{setlur2024rewardingprogress}, while
\citet{luo2025agentlightning} and \citet{peng2026hiper} assign credit to agent
transitions or hierarchical decisions.
\citet{team2026mirothinker} applies local
and global verification during inference. VerMem instead uses
operation-level and trajectory-level verification signals to train
the memory operation policy.

\section{Method}

\begin{figure*}[t]
    \centering
    \includegraphics[width=0.95\textwidth]
        {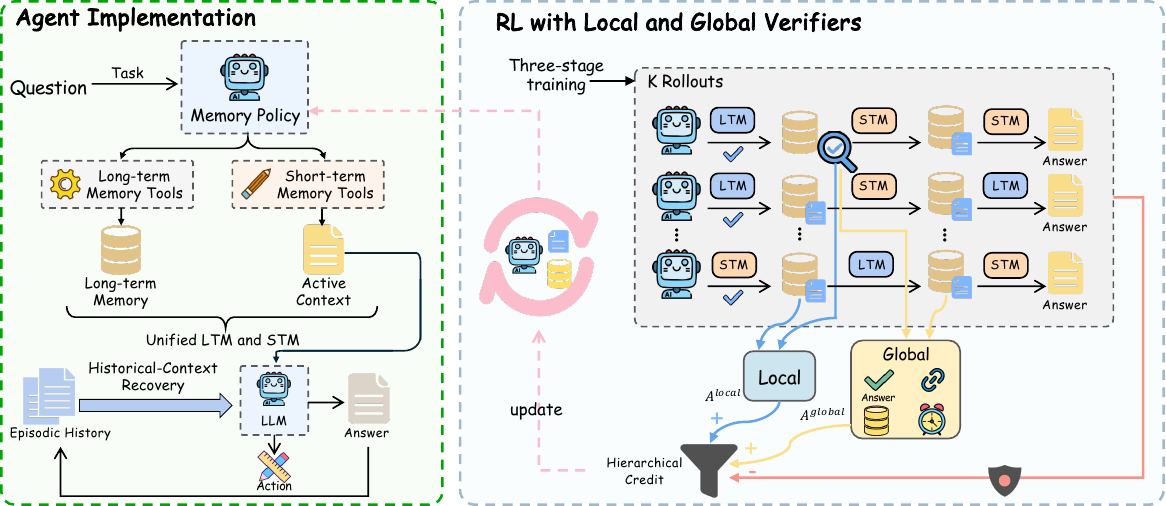}
   \caption{
Overview of VerMem. Left: the memory operation policy coordinates long-term and short-term memory tools to maintain long-term memory and the active context. Historical-context recovery restores relevant episodic history for the LLM, whose actions and answers are recorded for subsequent steps. Right: a three-stage curriculum generates \(K\) stateful rollouts. The local verifier evaluates executable memory transitions, while the global verifier assesses evidence coherence and terminal-memory consistency after task completion. These scores contribute to separately normalized local and global advantages; constraint costs remain separate. The verifier branch is used only during training.
}
    \label{fig:VerMem_overview}
\end{figure*}

We propose Verifiable Memory (VerMem), a unified framework
that maintains persistent LTM, a budgeted active context,
and episodic history as distinct states. As shown in
Figure~\ref{fig:VerMem_overview}, one memory operation
policy coordinates LTM maintenance, STM control, and
historical-context recovery. The policy is initialized with
SFT and optimized through a three-stage reinforcement
learning curriculum. The local verifier supplies operation-level
semantic scores, while the composite global branch supplies
trajectory-level credit from programmatic and global-verifier
components. Both verifiers are removed during inference.

\subsection{Problem Formulation}

\textbf{Unified memory operation policy formulation.}
The fixed task specification is denoted by \(q\). At memory-decision
step \(t\), the long-term memory \(M_t\), active context \(C_t\), and
same-task episodic history \(H_t\) define the conceptual state and its
policy-visible serialization:
\begin{equation}
\begin{aligned}
s_t &= (q,M_t,C_t,H_t), \\
x_t &= g(s_t).
\end{aligned}
\label{eq:state_serialization}
\end{equation}
Here, \(M_t\) stores persistent information for reuse in later steps or
stages, \(C_t\) is the budgeted context visible to the task solver, and
\(H_t\) preserves the ordered record of observations, task actions,
tool outputs, intermediate conclusions, and memory decisions from the
same task. Separating \(C_t\) from \(H_t\) allows information to leave
the active context without being removed from that record. The function
\(g\) retains the task specification and current task state, then
serializes budgeted views of \(M_t\), \(C_t\), and same-task \(H_t\)
within the 8,192-token policy-state limit used by the protocol. It does
not concatenate \(H_t\) without bound. Equation~\eqref{eq:state_serialization}
therefore distinguishes the conceptual state from the bounded input
presented to the policy.

Let \(v_t\) denote an operation type and \(\xi_t\) its structured
arguments. The complete generated memory command is
\(u_t=(v_t,\xi_t)\), where
\(v_t\in\mathcal V\cup\{\varnothing\}\), and \(\mathcal V\) denotes
the seven operation types formalized in
Equation~\eqref{eq:operation_sets}. The null command has empty
arguments. The complete command and memory transition are
\begin{equation}
\begin{aligned}
u_t &= (v_t,\xi_t), \\
u_t &\sim \pi_\theta(\cdot\mid x_t), \\
\widetilde s_t &= \mathcal T_m(s_t,u_t).
\end{aligned}
\label{eq:memory_command_transition}
\end{equation}
For a valid non-null command, \(\mathcal T_m\) commits the specified
memory transition. For \(v_t=\varnothing\), \(M_t\) and \(C_t\) are
unchanged,and the memory-state transition is an identity transition,
so \(\tilde{s}_t=s_t\). The identity decision is retained in
the rollout record but does not create a memory-state modification. The null
command remains a trainable decision in the rollout record. Rejected
non-null commands leave \(M_t\) and \(C_t\) unchanged and record the failed
decision and its reason in \(H_t\). They also incur a constraint cost. Thus,
Equation~\eqref{eq:memory_command_transition} defines \(\widetilde s_t\)
on every branch.

After the memory transition, the task solver acts on the resulting
active context, the environment returns an observation, and a separate
interaction transition constructs the next memory-decision state:
\begin{equation}
\begin{aligned}
b_t &\sim \pi_\phi^{\mathrm{task}}
(\cdot\mid q,\widetilde C_t),
&
z_{t+1} &\sim \mathcal P_{\mathrm{env}}(\cdot\mid b_t),
\\
s_{t+1} &= \mathcal T_e(\widetilde s_t,b_t,z_{t+1}).
\end{aligned}
\label{eq:environment_transition}
\end{equation}
The transition \(\mathcal T_e\) appends the task action and resulting
observation to \(H_t\) and constructs the next active context. The
memory transition in Equation~\eqref{eq:memory_command_transition} and
the task/environment transition in
Equation~\eqref{eq:environment_transition} have distinct roles.
Repeating them produces a task trajectory \(\tau\). Unified
management shares the state and task objective but does not merge LTM,
active context, and episodic history into one storage structure.

\textbf{Progressive training and hierarchical credit.}
The policy follows
\begin{equation}
\theta_{\mathrm{SFT}}
\rightarrow
\theta_A
\rightarrow
\theta_B
\rightarrow
\theta_C
=
\theta^\star.
\label{eq:parameter_curriculum}
\end{equation}
Phase A trains LTM maintenance, Phase B trains STM
control under distractors, and Phase C trains their
coordination in complete tasks. Equation~\eqref{eq:parameter_curriculum}
shows the parameter curriculum. For decision \(t\) in trajectory
\(k\), the local verifier produces a raw operation score for an
executable command, while the global branch combines post-trajectory
verifier scores with programmatic task, evidence-recall, and efficiency
signals. Normalization converts the two branches into
\(A_{t,k}^{\mathrm{local}}\) and \(A_k^{\mathrm{global}}\). The
executor separately records hard violations as \(c_{t,k}\). Their
credit is
\begin{equation}
A_{t,k}^{\mathrm{hier}}
=
\lambda_{\mathrm{local}}A_{t,k}^{\mathrm{local}}
+
\lambda_{\mathrm{global}}A_k^{\mathrm{global}}
-
\lambda_{\mathrm{constraint}}c_{t,k}.
\label{eq:hierarchical_credit}
\end{equation}
We set \(\lambda_{\mathrm{local}}=
\lambda_{\mathrm{global}}=\lambda_{\mathrm{constraint}}=1\).
The local term distinguishes executable commands within one
trajectory, while the global term preserves alignment with the
completed task. Equation~\eqref{eq:hierarchical_credit} keeps the local,
global, and constraint channels distinct, including for rejected
commands.

\subsection{Atomic Memory Tools}

The operation-type sets are
\begin{equation}
\begin{aligned}
\mathcal V_{\mathrm{LTM}}
&=\{\textsc{Add},\textsc{Update},
\\[-2pt]
&\hspace{1.2em}\textsc{Delete}\},\\
\mathcal V_{\mathrm{STM}}
&=\{\textsc{Retrieve},\textsc{Filter},
\\[-2pt]
&\hspace{1.2em}\textsc{SelectEpisode},\textsc{Summarize}\},\\
\mathcal V&=\mathcal V_{\mathrm{LTM}}\cup
\mathcal V_{\mathrm{STM}}.
\end{aligned}
\label{eq:operation_sets}
\end{equation}
VerMem exposes these seven atomic tools, summarized in
Table~\ref{tab:atomic_memory_tools} and formalized in
Equation~\eqref{eq:operation_sets}, together with the null action
\(\varnothing\). The LTM tools add, revise, or soft-delete
persistent information. The STM tools retrieve LTM content,
filter or summarize the active context, and restore relevant
same-task fragments from \(H_t\) through SelectEpisode.
The unified action space lets the policy choose between
persistent-memory maintenance and active-context control
from the current state. Detailed schemas, preconditions, and
transition checks are given in Supplementary Material, Section~A.

\begin{table}[t]
\centering
\small
\setlength{\tabcolsep}{3pt}
\begin{tabular}{@{}lll@{}}
\toprule
\textbf{Tool} & \textbf{Target} & \textbf{Function} \\
\midrule
Add
& LTM
& Add to \(M_t\) \\
Update
& LTM
& Create revised versions in \(M_t\) \\
Delete
& LTM
& Soft-delete entries in \(M_t\) \\
\midrule
Retrieve
& STM
& Retrieve \(M_t \rightarrow C_t\) \\
Filter
& STM
& Filter \(C_t\) \\
SelectEpisode
& STM
& Restore \(H_t \rightarrow C_t\) \\
Summarize
& STM
& Summarize \(C_t\) \\
\bottomrule
\end{tabular}
\caption{Atomic memory tools in VerMem.}
\label{tab:atomic_memory_tools}
\end{table}

\subsection{Training Pipeline}

\textbf{Curriculum and SFT warmup.}
VerMem uses an SFT warmup followed by three
reinforcement learning phases. The supervised set
\(\mathcal D_{\mathrm{SFT}}
=\{(x_i,v_i^\star,\xi_i^\star)\}_{i=1}^{N}\)
contains validated targets for the seven tools and
\(\varnothing\), where \(\xi_i^\star\) denotes structured
arguments and \(u_i^\star=(v_i^\star,\xi_i^\star)\). The objective is
\begin{equation}
\mathcal L_{\mathrm{SFT}}
=
-
\mathbb E_{(x_i,v_i^\star,\xi_i^\star)
\sim\mathcal D_{\mathrm{SFT}}}
\left[
\log
\pi_\theta
(v_i^\star,\xi_i^\star\mid x_i)
\right].
\label{eq:sft_objective}
\end{equation}
Equation~\eqref{eq:sft_objective} trains both operation selection and
structured argument generation.
Boundary templates and candidates from a frozen
DeepSeek-V3.2~\citep{liu2025deepseek} teacher are retained
only when they satisfy the tool schema and transition
constraints. The warmup teaches tool selection, argument
generation, and valid structured outputs, but does not
determine the delayed task utility of a memory decision.
Data construction, filtering, and coverage are detailed in
Supplementary Material, Section~D.

\textbf{Phase-wise reinforcement learning.}
Phase A optimizes
\(\mathcal V_{\mathrm{LTM}}\cup\{\varnothing\}\) on
construction, revision, deletion, and no-change decisions.
Phase B optimizes
\(\mathcal V_{\mathrm{STM}}\cup\{\varnothing\}\) on
retrieval, filtering, summarization, and historical-context
recovery under distractors. Phase C enables
\(\mathcal V\cup\{\varnothing\}\) in complete multi-step
tasks, where LTM and STM operations may alternate. Only
memory operation policy parameters are transferred across phases; \(M_t\),
\(C_t\), and \(H_t\) are initialized independently for every
training instance.

\textbf{Stateful trajectory collection.}
At policy update \(n\) of phase \(p\in\{A,B,C\}\), the current
policy samples \(K\) trajectories from the same
phase-specific initial state:
\begin{equation}
\begin{aligned}
\tau_{k,p}^{(q,n)}
&\sim
\mathbb P_{p,n}
\left(\tau\mid s_{0,p}^{(q)}\right),
\quad k=1,\ldots,K,
\\
\mathcal G_{q,p}^{(n)}
&=
\{\tau_{1,p}^{(q,n)},\ldots,\tau_{K,p}^{(q,n)}\}.
\end{aligned}
\label{eq:stateful_rollout}
\end{equation}
Each rollout stores both the post-memory state \(\widetilde s_t\) and
the post-interaction state \(s_{t+1}\), so that memory-transition credit
is not conflated with the subsequent task action. The candidate group
in Equation~\eqref{eq:stateful_rollout} supports subsequent local and
global credit assignment.
Phase-specific states, termination rules, and token masks
are provided in Supplementary Material, Sections~D and~F.

\subsection{Local and Global Verifier-Guided Hierarchical
Credit Assignment}

We optimize the memory operation policy with local and global credit
signals. For each task \(q\), the \(K\) trajectories sampled from the
same initial state form the candidate group
\(\mathcal G_q=\{\tau_1^{(q)},\ldots,\tau_K^{(q)}\}\). Let
\(\mathcal B_{\mathrm{op}}\) contain every memory decision collected in
the current update, including null and rejected commands, and let
\(\mathcal B_{\mathrm{exec}}\subseteq\mathcal B_{\mathrm{op}}\)
contain valid non-null commands and valid null decisions. Structural,
safety, and budget violations are recorded separately as \(c_{t,k}\).
For \((q,k,t)\in\mathcal B_{\mathrm{exec}}\), the local verifier
produces \(r_{t,k}^{\mathrm{local}}\), which evaluates task relevance,
evidence grounding, local progress, and information fidelity. The null
decision is treated as a valid identity command and is scored with its
own rubric. Rejected commands are not sent to the local verifier; their
local advantage is set to zero. After task termination, the global
trajectory branch produces the composite score
\(r_k^{\mathrm{global}}\). Its verifier and programmatic components are
defined below.

The composite global score is normalized within the candidate group
for the same task. Local scores are normalized by operation type. The
decision subset and two advantages are
\begin{equation}
\begin{aligned}
\mathcal B_v
&=
\{(q,k,t)\in\mathcal B_{\mathrm{exec}}:v_{t,k}=v\},
\\
A_k^{\mathrm{global}}
&=
\frac{r_k^{\mathrm{global}}-\mu_q^{\mathrm{global}}}
{\sigma_q^{\mathrm{global}}+\epsilon},
\\
A_{t,k}^{\mathrm{local}}
&=
\frac{r_{t,k}^{\mathrm{local}}-
\mu_{v_{t,k}}^{\mathrm{local}}}
{\sigma_{v_{t,k}}^{\mathrm{local}}+\epsilon}.
\end{aligned}
\label{eq:normalized_advantages}
\end{equation}
The global statistics are computed from the \(K\)
trajectories in \(\mathcal G_q\), while local statistics are
computed from operations of the same type in the current
update batch. This operation-wise normalization reduces
scale differences across tool-specific verifier rubrics. When
an update contains too few instances of an operation type,
phase-specific running statistics are used. The corresponding
thresholds and update rules are provided in Supplementary Material,
Section~F. The local expression in
Equation~\eqref{eq:normalized_advantages} is defined only for decisions
in \(\mathcal B_{\mathrm{exec}}\). For rejected commands, it is not
evaluated and \(A_{t,k}^{\mathrm{local}}=0\).

At each memory decision, Equation~\eqref{eq:hierarchical_credit}
combines the two advantages with the constraint cost.
The local advantage is assigned only to the current atomic
operation, while the global advantage provides task-level
supervision to all operations in the trajectory. Different
memory decisions within one trajectory can therefore
receive different learning signals. Poorly rated operations in a
successful trajectory therefore need not receive the trajectory's full
positive credit, while useful operations in an unsuccessful trajectory
may retain a positive local contribution. The final sign still depends
on the weighted sum of the local, global, and constraint terms.
Rejected commands are not evaluated by the local semantic verifier, so
\(A_{t,k}^{\mathrm{local}}=0\). They still receive the trajectory-level
\(A_k^{\mathrm{global}}\), while \(c_{t,k}\) provides their direct
operation-level penalty.

Let \(y_{t,k,1:L_{t,k}}\) be the generated tokens that serialize the
complete command \(u_{t,k}\), including its structured arguments, and
let \(h_{t,k,j}=(x_{t,k},y_{t,k,<j})\) be the token prefix. The
token-level policy ratio and KL term are
\begin{equation}
\begin{aligned}
\rho_{t,k,j}(\theta)
&=
\frac{\pi_\theta(y_{t,k,j}\mid h_{t,k,j})}
{\pi_{\theta_{\mathrm{old}}}(y_{t,k,j}\mid h_{t,k,j})},
\\
d_{t,k,j}^{\mathrm{KL}}
&=D_{\mathrm{KL}}\!\left(
\pi_\theta(\cdot\mid h_{t,k,j})
\,\|\,
\pi_{\mathrm{ref}}(\cdot\mid h_{t,k,j})
\right).
\end{aligned}
\label{eq:token_ratio_kl}
\end{equation}
Here, \(\theta_{\mathrm{old}}\) denotes the policy frozen
before the current update, and \(\pi_{\mathrm{ref}}\) is the
phase-specific reference policy. The coefficient
\(\epsilon_{\mathrm{clip}}\) controls PPO clipping, while
\(\beta_{\mathrm{KL}}\) weights the KL regularization term. All command tokens share \(A_{t,k}^{\mathrm{hier}}\), but the clipped
ratio and KL term in Equation~\eqref{eq:token_ratio_kl} are evaluated
token by token. The operation-normalized objective is
\begin{equation}
\begin{aligned}
\ell_{t,k,j}(\theta)
={}&\min\big(
\rho_{t,k,j}(\theta)A_{t,k}^{\mathrm{hier}},
\\[-2pt]
&\operatorname{clip}(\rho_{t,k,j}(\theta),
1-\epsilon_{\mathrm{clip}},1+\epsilon_{\mathrm{clip}})
A_{t,k}^{\mathrm{hier}}
\big)
\\[-2pt]
&-\beta_{\mathrm{KL}}d_{t,k,j}^{\mathrm{KL}},
\\
\mathcal J_{\mathrm{GRPO}}(\theta)
={}&\frac{1}{|\mathcal B_{\mathrm{op}}|}
\sum_{(q,k,t)\in\mathcal B_{\mathrm{op}}}
\frac{1}{L_{t,k}}
\sum_{j=1}^{L_{t,k}}\ell_{t,k,j}(\theta).
\end{aligned}
\label{eq:grpo_objective}
\end{equation}
Equation~\eqref{eq:grpo_objective} uses the clipped surrogate from PPO
and candidate-group normalization from GRPO
\citep{schulman2017proximal,shao2024deepseekmath}. The memory operation policy loss is
applied only to tokens that serialize the atomic memory command.
Task-solver outputs and environment observations are excluded from this
loss. Task specifications, state-serialization tokens, and padding tokens
are also excluded. Operations at different time steps retain different
advantages, and the factor \(1/L_{t,k}\) prevents commands with longer
arguments from dominating the update. Loss masks and stabilization
settings are provided in Supplementary Material, Sections~B and~F. The
local verifier, global verifier, and reward computation are used only
during training.

\subsection{Reward Function Design}

The training signal has three explicitly separated channels: semantic
quality for executable memory commands, global utility for the completed
trajectory, and hard execution constraints. Let \(T_k\) be the number
of memory-decision opportunities in \(\tau_k\), including null and
rejected commands, and let \(\mathcal E_k\) index its executable
commands, including valid null decisions. Null decisions count toward
the decision horizon but not as memory-tool calls. Keeping the three
channels separate prevents a hard rejection from also receiving an
undefined or duplicated semantic penalty.

\textbf{Local operation reward.}
The local component is
\begin{equation}
R^{\mathrm{local}}(\tau_k)
=
\frac{1}{T_k}
\sum_{t\in\mathcal E_k}
r_{t,k}^{\mathrm{local}}.
\label{eq:local_reward}
\end{equation}
Equation~\eqref{eq:local_reward} normalizes by all memory-decision
opportunities while summing only scores produced for executable
decisions. It is therefore zero when \(\mathcal E_k\) is empty and is used
for analysis and monitoring, not as a substitute for the operation-wise
advantages.
The score evaluates only the current atomic memory
operation and its realized state transition. It measures task
relevance, evidence grounding, local progress, and
information fidelity. For LTM operations, it evaluates
whether information has persistent value, whether an update
agrees with new evidence, and whether a deletion is
justified. For STM operations, it evaluates whether
retrieval or episode-level context selection supplies missing
information and whether filtering or summarization preserves
useful evidence while reducing context load. Structural
validity is excluded from this semantic score and is handled
by the auxiliary channel.

\textbf{Composite global trajectory reward.}
The composite global branch is computed after task termination:
\begin{equation}
\begin{aligned}
r_k^{\mathrm{global}}
={}&
\tfrac14\big(
r_k^{\mathrm{task}}+r_k^{\mathrm{evid}}
+r_k^{\mathrm{state}}+r_k^{\mathrm{eff}}
\big).
\end{aligned}
\label{eq:global_reward}
\end{equation}
All four terms lie in \([0,1]\). The task score
\(r_k^{\mathrm{task}}\) and annotated supporting-fact recall
\(r_k^{\mathrm{sup}}\) are computed programmatically. The global
verifier returns only an evidence-coherence score
\(v_k^{\mathrm{coh}}\) and a terminal-memory-consistency score
\(v_k^{\mathrm{state}}\). The evidence and state components are
\begin{equation}
\begin{aligned}
r_k^{\mathrm{evid}}
&=\tfrac12\big(r_k^{\mathrm{sup}}+v_k^{\mathrm{coh}}\big),
\\
r_k^{\mathrm{state}}
&=v_k^{\mathrm{state}}.
\end{aligned}
\label{eq:evidence_state_rewards}
\end{equation}
Equations~\eqref{eq:global_reward} and
\eqref{eq:evidence_state_rewards} make the boundary explicit: the global
verifier supplies only \(v_k^{\mathrm{coh}}\) and
\(v_k^{\mathrm{state}}\), while the composite global branch also includes
programmatic task correctness, supporting-fact recall, and efficiency.
Terminal-memory consistency requires current active LTM and context to
agree with observed evidence and treats \(H_t\) as an ordered
provenance record: superseded or failed events may remain in history if
their status is explicit. Since policy training uses only HotpotQA,
\(r_k^{\mathrm{task}}\) is derived from answer correctness.
Environment success, planning progress, and goal-state satisfaction are
used as evaluation metrics on the downstream agent benchmarks rather
than as policy-training rewards.

Let
\(\overline C_k^{\mathrm{online}}\in[0,1]\) denote the
programmatically computed online cost from tokens, task steps, and
executed non-null memory-tool calls. These components are separately
min--max normalized within the \(K\) candidates for the same task and
then averaged. The efficiency term is
\begin{equation}
r_k^{\mathrm{eff}}
=
\mathbb I
\left[
r_k^{\mathrm{task}}
\geq
\delta
\right]
\left(
1-
\overline C_k^{\mathrm{online}}
\right).
\label{eq:efficiency_reward}
\end{equation}
We use \(\delta=0.5\). This gate prevents low-cost failures from
receiving an efficiency reward in Equation~\eqref{eq:efficiency_reward}.

\textbf{Auxiliary constraints.}
The auxiliary cost is
\begin{equation}
P_{\mathrm{aux}}(\tau_k)
=
\frac{1}{T_k}
\sum_{t=1}^{T_k}
c_{t,k}.
\label{eq:auxiliary_cost}
\end{equation}
Equation~\eqref{eq:auxiliary_cost} averages direct constraint costs over
the memory-decision horizon.
The step cost records discrete or hard violations, including
invalid tools or arguments, failed state transitions,
unsupported writes or updates, unjustified deletions,
making protected evidence operationally inaccessible within the
remaining decision and token budget, context overflow,
exceeding the tool or interaction budget, and information
leakage in no-reference settings. Routine token, step, and
tool costs do not enter this channel; they are evaluated
continuously by the global efficiency term. Terminal budget violations
are attached to the final memory decision that precedes termination.

The scalar displayed in the All-Returns training curve is used only for
monitoring:
\begin{equation}
m_k^{\mathrm{all}}
=\operatorname{clip}\!\left(
\frac{R^{\mathrm{local}}(\tau_k)+r_k^{\mathrm{global}}
-P_{\mathrm{aux}}(\tau_k)}{2},0,1\right).
\label{eq:all_returns_monitor}
\end{equation}
The Answer-Only curve analogously displays
\begin{equation}
m_k^{\mathrm{ans}}
=\operatorname{clip}\!\left(
r_k^{\mathrm{task}}-P_{\mathrm{aux}}(\tau_k),0,1\right).
\label{eq:answer_only_monitor}
\end{equation}
Equations~\eqref{eq:all_returns_monitor} and
\eqref{eq:answer_only_monitor} are strategy-specific monitoring
values. They are not policy advantages and are not directly comparable in
absolute level. During policy optimization, the local and global
signals are normalized separately, while \(c_{t,k}\) is assigned
directly to the corresponding decision. Tool-specific rubrics,
normalization details, completion thresholds, and constraint triggers
are provided in Supplementary Material, Section~F.

\section{Experiments}

\subsection{Experimental Setup}

\textbf{Datasets.}
We evaluate VerMem on
ALFWorld~\citep{shridhar2020alfworld},
SciWorld~\citep{wang2022scienceworld},
PDDL~\citep{ma2024agentboard},
BabyAI~\citep{chevalierboisvert2018babyai}, and
HotpotQA~\citep{yang2018hotpotqa}. VerMem is fine-tuned
only on the HotpotQA training split, whose supporting facts
and distractors support LTM maintenance, STM control, and
historical-context recovery, and is then evaluated directly
on all five benchmarks. Memory states are isolated by task
instance. Each episode starts with empty \(M_0\) and
\(H_0\), while \(C_0\) contains only the task instruction
and initial input.

\textbf{Evaluation Metrics.}
We report Success Rate (successful episodes divided by evaluated
episodes) on ALFWorld, SciWorld, and BabyAI, Progress Rate on PDDL,
and the Qwen-Max LLM-as-a-Judge score \(J_{\mathrm{judge}}\) on
HotpotQA. The main table reports \(100J_{\mathrm{judge}}\). Qwen-Max is
used only as a held-out evaluator; it is not used for policy training,
verifier scoring, or checkpoint selection. Let \(S^*\) be a target
macro-average Success Rate and \(B^*\) an online-token budget.
\(\mathrm{TN}@S^*\) is the linearly interpolated online-token budget at
which the mean success curve first reaches the target \(S^*\), using the
two adjacent evaluated budget points that bracket the target.
\(\mathrm{SR}@B^*\) is the macro-average Success Rate at \(B^*\). The
reward ablation additionally reports
Memory Quality (MQ), average online token number (TN), and average
memory-tool calls (TC).

\textbf{Baselines and Backbones.}
Base uses the common task solver without an explicit memory operation
policy or external persistent-memory store. We compare it with
LangMem~\citep{langchain2025langmem}, A-Mem~\citep{xu2025amem}, Mem0 and
its graph-based variant Mem0\(^{g}\)~\citep{chhikara2025mem0}, and
AgeMem~\citep{yu2026agemem} under Qwen2.5-7B-Instruct
\citep{Qwen2024Qwen2} and Qwen3-4B-Instruct
\citep{yang2025qwen3}.
VerMem-noVerify is a matched internal control that
removes semantic feedback from the local and global verifiers while
retaining the same action space, SFT initialization, curriculum,
task-outcome reward, transition constraints, and GRPO configuration.
Complete implementation and baseline settings are provided in
Supplementary Material, Sections~C and~F.

\subsection{Main Results}

\begin{table*}[t]
\centering
\setlength{\tabcolsep}{4pt}
\renewcommand{\arraystretch}{1.02}

\begin{tabular}{@{}lcccccc@{}}
\toprule

\textbf{Method}
&
\textbf{ALFWorld}
&
\textbf{SciWorld}
&
\textbf{PDDL}
&
\textbf{BabyAI}
&
\textbf{HotpotQA \(100J_{\mathrm{judge}}\)}
&
\textbf{Average}
\\

\midrule

\multicolumn{7}{@{}l}{\textbf{Qwen2.5-7B-Instruct}}\\
Base
& 27.16
& 13.80
& 10.15
& 50.80
& 38.36
& 28.05
\\

LangMem
& 38.27
& 28.29
& 15.85
& 51.34
& 37.43
& 34.24
\\

A-Mem
& 34.68
& 28.06
& 18.39
& 58.82
& 43.95
& 36.78
\\

Mem0
& 37.49
& 26.99
& 13.96
& 60.58
& 46.66
& 37.14
\\

Mem0\(^{g}\)
& 35.34
& 30.50
& 14.86
& 58.78
& 42.06
& 36.31
\\

AgeMem
& \secondscore{41.07}
& \secondscore{35.55}
& 17.31
& \secondscore{61.42}
& \secondscore{54.44}
& \secondscore{41.96}
\\

VerMem-noVerify
& 40.18
& 33.94
& \bestscore{22.41}
& 59.21
& 53.08
& 41.76
\\

\textbf{VerMem (Ours)}
& \bestscore{46.30}
& \bestscore{43.39}
& \secondscore{22.06}
& \bestscore{65.55}
& \bestscore{62.75}
& \bestscore{48.01}
\\

\midrule

\multicolumn{7}{@{}l}{\textbf{Qwen3-4B-Instruct}}\\
Base
& 38.51
& 47.89
& 30.14
& 55.83
& 47.48
& 43.97
\\

LangMem
& 40.89
& 50.42
& 28.42
& 53.80
& 42.70
& 43.25
\\

A-Mem
& 34.31
& 50.14
& 34.41
& 61.35
& 48.48
& 45.74
\\

Mem0
& 41.17
& 51.38
& 31.72
& 60.05
& 39.16
& 44.70
\\

Mem0\(^{g}\)
& 36.69
& 47.76
& 29.61
& 57.59
& 38.12
& 41.95
\\

AgeMem
& \secondscore{48.97}
& \secondscore{59.48}
& \secondscore{35.07}
& \secondscore{72.56}
& \secondscore{55.49}
& \secondscore{54.31}
\\

VerMem-noVerify
& 48.63
& 58.92
& 34.76
& 72.11
& 53.79
& 53.64
\\

\textbf{VerMem (Ours)}
& \bestscore{53.73}
& \bestscore{66.37}
& \bestscore{38.75}
& \bestscore{76.77}
& \bestscore{63.61}
& \bestscore{59.85}
\\

\bottomrule
\end{tabular}

\caption{
Main results under two LLM backbones. We report Success
Rate on ALFWorld, SciWorld, and BabyAI, Progress Rate
on PDDL, and \(100J_{\mathrm{judge}}\) on HotpotQA. Each entry
is the mean over three independent runs with seeds 42, 43, and 44.
Average is the unweighted arithmetic mean of the five displayed scores.
The best result in each backbone setting is shown in bold, and the
second-best result is underlined.
}
\label{tab:main_results}
\end{table*}

\textbf{Overall comparison.}
Table~\ref{tab:main_results} reports the main results under
both backbones. With Qwen2.5-7B-Instruct, VerMem
obtains an average score of \(48.01\). This is \(19.96\)
points above Base, corresponding to a relative gain of
\(71.16\%\), and \(6.05\) points above the strongest
external method, AgeMem. VerMem ranks first on
ALFWorld, SciWorld, BabyAI, HotpotQA, and Average.
PDDL is the only exception, where VerMem-noVerify is
\(0.35\) points higher. VerMem still exceeds the strongest
external PDDL baseline by \(3.67\) points. With
Qwen3-4B-Instruct, VerMem ranks first in all six reported
columns and reaches an average score of \(59.85\),
outperforming AgeMem by \(5.54\) points. The consistent
improvement is observed with both evaluated backbones rather than only
one of them.

\textbf{Performance across task types.}
The improvements vary with the memory requirements of
each task. With Qwen2.5-7B-Instruct, VerMem exceeds the
strongest external baseline by \(5.23\), \(7.84\), \(3.67\),
\(4.13\), and \(8.31\) points on ALFWorld, SciWorld,
PDDL, BabyAI, and HotpotQA. The corresponding gains
with Qwen3-4B-Instruct are \(4.76\), \(6.89\), \(3.68\),
\(4.21\), and \(8.12\) points. The largest gains occur on HotpotQA and
SciWorld. This pattern is consistent with the greater need for
multi-hop evidence aggregation and long interaction-state tracking in
these tasks. The smaller gain on PDDL is consistent with its more
structured state representation and feedback, although the present
experiments do not isolate this factor.

\textbf{Cross-task transfer.}
VerMem is fine-tuned only on HotpotQA, yet it consistently
improves over the strongest baselines on ALFWorld,
SciWorld, PDDL, and BabyAI. The gains range from \(3.67\)
to \(7.84\) points with Qwen2.5-7B-Instruct and from
\(3.68\) to \(6.89\) points with Qwen3-4B-Instruct. These
environments require environment tracking,
scientific-progress maintenance, planning-state control, and
instruction execution. The transfer results indicate that the learned
memory operation policy captures reusable
memory behavior rather than task-specific document patterns.

\begin{figure}[t]
\centering
\includegraphics[width=\columnwidth]
{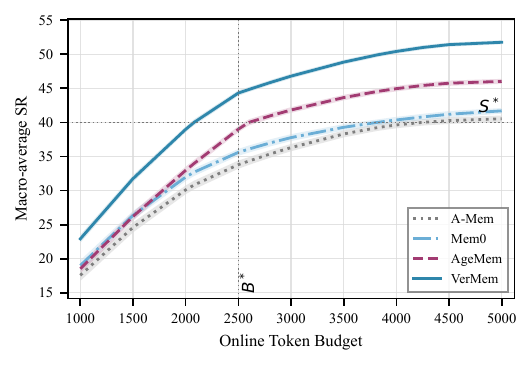}
\caption{
Efficiency--performance frontier under different online
token budgets on ALFWorld, SciWorld, and BabyAI using Qwen2.5-7B-Instruct. Curves compare A-Mem, Mem0, AgeMem, and VerMem, and report macro-average Success Rate over three random seeds. Shaded regions denote the standard deviation across seeds. The horizontal and vertical
dotted lines mark \(S^*=40.0\) and \(B^*=2{,}500\), respectively.
}
\label{fig:efficiency_frontier}
\end{figure}

\begin{table}[t]
\centering
\begin{tabular}{lcc}
\toprule
\textbf{Method}
&
\(\mathbf{\mathrm{TN}@S^*}\downarrow\)
&
\(\mathbf{\mathrm{SR}@B^*}\uparrow\)
\\
\midrule
A-Mem               & 4,250 & 33.80 \\
Mem0                & 3,820 & 35.60 \\
AgeMem              & 2,600 & 39.00 \\
\textbf{VerMem}     & \textbf{2,080} & \textbf{44.30} \\
\bottomrule
\end{tabular}
\caption{
Efficiency--performance comparison under controlled
online-token budgets using Qwen2.5-7B-Instruct, with \(S^*=40.0\) and
\(B^*=2{,}500\). \(\mathrm{TN}@S^*\) is linearly interpolated from the
two adjacent evaluated budget points that bracket \(S^*\). Lower
\(\mathrm{TN}@S^*\) and higher \(\mathrm{SR}@B^*\) are better.
}
\label{tab:efficiency_results}
\end{table}

\textbf{Efficiency--performance frontier.}
Figure~\ref{fig:efficiency_frontier} and
Table~\ref{tab:efficiency_results} show that VerMem
maintains the highest macro-average SR across the evaluated
budgets. Its mean curve first reaches \(S^*=40.0\) at the linearly
interpolated estimate of \(2{,}080\) online tokens, reducing
\(\mathrm{TN}@S^*\) by \(20.0\%\) relative
to AgeMem, and achieves an \(\mathrm{SR}@B^*\) of \(44.30\)
at \(B^*=2{,}500\), exceeding AgeMem by \(5.30\) points.
Both verifiers are disabled during evaluation, so these measurements do
not include inference-time verification calls.

\subsection{Ablation Studies}

\begin{figure*}[t]
\centering
\includegraphics[width=0.98\textwidth]
{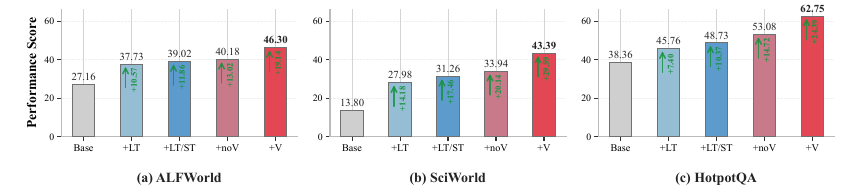}
\caption{
Cumulative component ablation with
Qwen2.5-7B-Instruct. ALFWorld and SciWorld report
Success Rate. HotpotQA reports \(100J_{\mathrm{judge}}\). Base has no
explicit memory operation policy. +LT adds LTM operations. +LT/ST adds the full
LTM/STM tool space. +noV adds stateful RL without
semantic verifier feedback. +V adds the local and global credit
branches.
}
\label{fig:component_ablation}
\end{figure*}

\textbf{Unified memory components.}
Figure~\ref{fig:component_ablation} shows cumulative gains
from LTM operations, the complete LTM/STM action space,
stateful RL, and verifier-guided credit. LTM provides the largest
initial improvement, STM control adds further gains on
SciWorld and HotpotQA, and full VerMem improves over
+noV by \(6.12\), \(9.45\), and \(9.67\) points. The
progression is consistent with contributions from both the unified
action space and verifier-guided optimization.

\begin{table}[t]
\centering
\setlength{\tabcolsep}{3pt}
\renewcommand{\arraystretch}{0.92}
\resizebox{\columnwidth}{!}{%
\begin{tabular}{lccc}
\toprule
\textbf{Variant}
& \textbf{ALFWorld SR}
& \textbf{SciWorld SR}
& \textbf{HotpotQA \(100J_{\mathrm{judge}}\)} \\
\midrule
VerMem-noVerify
& 40.18 & 33.94 & 53.08 \\
+Local
& 44.72 & 39.81 & 58.26 \\
+Global
& 43.31 & 37.96 & 60.18 \\
\textbf{+Local+Global}
& \textbf{46.30}
& \textbf{43.39}
& \textbf{62.75} \\
\bottomrule
\end{tabular}
}
\caption{Ablation of the local and global credit branches. VerMem-noVerify
uses a normalized task-outcome advantage and constraints. +Local adds
the local operation advantage. +Global replaces the task-only
trajectory advantage with the composite global trajectory advantage.
Local+Global uses both the local and composite global advantages and
denotes full VerMem. Architecture, SFT initialization, training data,
constraints, and GRPO settings are otherwise shared.}
\label{tab:verifier_ablation}
\end{table}

\textbf{Local and global credit branches.}
Table~\ref{tab:verifier_ablation} shows complementary
effects. Local feedback yields larger standalone gains on ALFWorld and
SciWorld, whereas the global branch yields a larger standalone gain on
HotpotQA. This pattern is consistent with, but does not by itself prove,
the intended distinction between operation-level and trajectory-level
credit. Their combination performs best on all three evaluated tasks.

\begin{figure}[t]
\centering
\includegraphics[width=\columnwidth]
{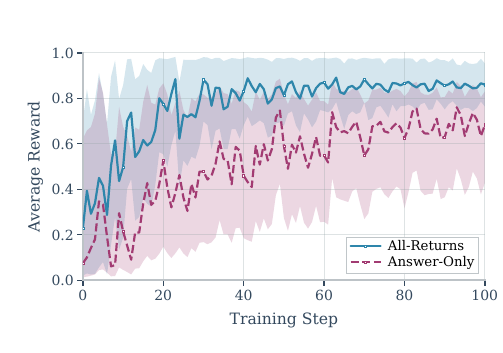}
\caption{
GRPO training-reward convergence on HotpotQA using
Qwen2.5-7B-Instruct. Training Step indexes the 101 recorded training
points over the 3,500-update GRPO schedule, with adjacent points
separated by 35 policy updates. Curves show the unsmoothed mean over
three runs, and shaded regions denote one standard deviation.
Answer-Only and All-Returns use the monitoring mappings defined in the
reward section; their absolute levels are not directly comparable
because the mappings differ.
}
\label{fig:reward_convergence}
\end{figure}

\begin{table}[t]
\centering
\begin{tabular}{lcccc}
\toprule
\textbf{Strategy}
&
\(\mathbf{J_{\mathrm{judge}}}\uparrow\)
&
\(\mathbf{TN}\downarrow\)
&
\(\mathbf{MQ}\uparrow\)
&
\textbf{TC}
\\
\midrule
Answer-Only
& 0.531
& 2,352
& 0.491
& 4.47
\\

\textbf{All-Returns}
& \textbf{0.628}
& \textbf{2,184}
& \textbf{0.673}
& 5.63
\\
\bottomrule
\end{tabular}
\caption{
Reward-function ablation on HotpotQA using
Qwen2.5-7B-Instruct. \(J_{\mathrm{judge}}\) denotes the
LLM-as-a-Judge score on \([0,1]\), TN is the average online token
number, MQ is Memory Quality, and TC is the average
number of memory-tool calls.
}
\label{tab:reward_ablation}
\end{table}

\textbf{Reward function.} The monitoring curves in
Figure~\ref{fig:reward_convergence} suggest earlier stabilization and
lower late-stage variability for All-Returns. Because the two
strategies use different monitoring definitions, their absolute reward
levels should not be compared directly. Table~\ref{tab:reward_ablation}
shows that All-Returns raises \(J_{\mathrm{judge}}\) from \(0.531\) to
\(0.628\) and MQ from \(0.491\) to \(0.673\), while reducing TN by
\(7.14\%\). TC increases from \(4.47\) to \(5.63\); together with the
higher \(J_{\mathrm{judge}}\) and MQ, this shows that the token reduction
is not obtained merely by suppressing all memory-tool calls.

\section{Limitations}

Our evaluation resets all memory states after each episode, so it
measures within-episode persistence and cross-benchmark transfer of the
learned policy rather than cross-session or cross-user memory retention.
The memory operation policy is trained on constructed HotpotQA states and
evaluated with two Qwen backbones; the findings may not transfer to
other model families or domains. HotpotQA \(J_{\mathrm{judge}}\) and MQ
use a fixed proprietary Qwen-Max snapshot, which limits fully
independent replication despite disclosure of the snapshot and prompts.
The frozen DeepSeek-V3.2 teacher and verifiers also introduce offline
training cost. Finally, learned verifier scores may inherit evaluator
errors, and the present study does not establish robustness under
adversarial, privacy-sensitive, or conflicting memory content.

\section{Conclusion}

We presented Verifiable Memory (VerMem), a unified
framework for learning LTM and STM management in
long-horizon LLM agents. VerMem maintains persistent
memory, active context, and episodic history as distinct
states, while one memory operation policy coordinates seven
atomic tools for persistent-memory maintenance,
active-context control, and historical-context recovery.
Training combines an SFT warmup with a three-stage
reinforcement learning curriculum. The local verifier
evaluates realized atomic memory transitions, while the
global verifier evaluates completed trajectories and terminal
memory states. Both verifiers are removed during inference.

Experiments across five benchmarks and two backbones show
that VerMem consistently outperforms strong memory
baselines and transfers from HotpotQA to interactive,
planning, and instruction-following tasks. The
budget-controlled evaluation further demonstrates a stronger
efficiency--performance trade-off under limited online
tokens. The ablation results confirm the contributions of
unified LTM/STM control, stateful reinforcement learning,
complementary verifier signals, and the complete reward
design. These findings show that coordinated memory
management and multi-granularity credit assignment provide
an effective basis for reliable long-horizon agent behavior.


\clearpage
\appendix

\section{Method and Tool Specifications}
\label{app:method_tools}

The following sections provide the runtime details omitted from the
main paper. We first describe how \(M_t\), \(C_t\), and
\(H_t\) are maintained during task execution. We then give
the tool-specific inputs, preconditions, and state-transition
rules of the seven atomic memory tools. The final two
subsections explain historical-context recovery and failure
handling. Training-instance construction, verifier rubrics,
and experimental configurations are provided in the
subsequent appendices.

\subsection{Runtime State and Execution Protocol}
\label{app:runtime_protocol}

VerMem represents the memory-decision state as
\(s_t=(q,M_t,C_t,H_t)\). The task specification \(q\)
remains fixed within one task instance. The long-term memory
\(M_t\) contains persistent entries that can be reused in
later steps or stages. Each entry records its content, source,
current status, and revision history. The active context
\(C_t\) contains the budgeted information currently visible
to the task solver. The episodic history \(H_t\) preserves the ordered
record of observations, task actions, tool outputs, intermediate
conclusions, and memory decisions from the same task.

The active context and episodic history serve different
purposes. Removing information from \(C_t\) reduces the
content visible to the task solver, but does not remove the
corresponding event from \(H_t\). This separation allows
earlier evidence to be recovered after the task focus changes.

At each memory-decision step, the policy generates a complete
command \(u_t=(v_t,\xi_t)\), where
\(v_t\in\mathcal V\cup\{\varnothing\}\). Each non-null command
specifies one atomic tool and its structured arguments. The policy
samples \(u_t\) from the bounded serialization
\(x_t=g(s_t)\) defined in Equation~\eqref{eq:state_serialization}.
The policy receives a bounded serialization of the task specification
and budgeted views of \(M_t\), \(C_t\), and \(H_t\). This policy-visible
serialization is distinct from the active-context budget governing
\(C_t\).
Before the command changes the real memory state, the proposed
transition is checked against the tool schema, target state,
required arguments, source information, and current budget.
The transition follows Equation~\eqref{eq:memory_command_transition}.
Each valid non-null command commits its realized memory transition. The
null command leaves \(M_t\) and \(C_t\) unchanged. On these states,
\(\mathcal T_m\) is the identity transition. The identity decision is
retained in the rollout record and does not create a memory-state
modification.
If a non-null command is rejected, \(M_t\) and \(C_t\) remain
unchanged, while the failed command and its reason are recorded in
\(H_t\). During training, the corresponding violation is recorded
through \(c_{t,k}\).

After the memory transition, the task solver continues with
the resulting active context. The task action and resulting observation
are then incorporated through the environment interaction transition
in Equation~\eqref{eq:environment_transition}, which constructs
\(s_{t+1}\). The following procedure summarizes this runtime process.
The local and global verifiers are used only
during training and do not participate in the reported
inference process.

\paragraph{Unified runtime procedure.}
Starting from \(M_0\), \(C_0\), and \(H_0\), each memory-decision
step performs the following operations:
\begin{enumerate}
    \item construct \(s_t=(q,M_t,C_t,H_t)\) and its bounded
    serialization \(x_t=g(s_t)\);
    \item sample the complete command
    \(u_t\sim\pi_\theta(\cdot\mid x_t)\);
    \item compute
    \(\widetilde s_t=\mathcal T_m(s_t,u_t)\) under the applicable
    execution branch. The valid non-null command commits its realized
    transition. The valid null command leaves \(M_t\) and \(C_t\)
    unchanged, creates no memory-state modification, and retains the
    identity decision in the rollout record. The rejected non-null
    command preserves \(M_t\) and \(C_t\), records the failed command
    and reason in \(H_t\), and receives the corresponding
    \(c_{t,k}\) during training;
    \item sample the task action \(b_t\) from the resulting active
    context, collect the next observation, and construct
    \(s_{t+1}=\mathcal T_e(\widetilde s_t,b_t,z_{t+1})\);
    \item repeat until task termination or a configured rollout limit.
\end{enumerate}

\subsection{Atomic Tool Schemas and State Transitions}
\label{app:tool_schemas}

Each non-null memory command selects one of the seven tools
defined in the main paper and supplies its required
arguments. Table~\ref{tab:app_atomic_tools} summarizes
the required information, state transition, and principal
validity condition of each tool. Optional implementation
fields and benchmark-specific budgets are reported with the
experimental configuration.

\begin{table*}[t]
\centering
\footnotesize
\setlength{\tabcolsep}{4pt}
\renewcommand{\arraystretch}{1.06}
\begin{tabular}{
@{}
p{0.20\textwidth}
p{0.46\textwidth}
p{0.26\textwidth}
@{}}
\toprule
\textbf{Tool and Required Information}
&
\textbf{State Transition}
&
\textbf{Key Validity Condition}
\\
\midrule

Add: new content and its source
&
Create a new persistent entry in \(M_t\).
&
The content is non-empty, supported, and not a duplicate
of an active entry.
\\

Update: target entry, revised content, and source
&
Create a revised version in \(M_t\) while retaining the
previous version.
&
The target exists, and the revision is supported by the
current evidence.
\\

Delete: target entry and deletion reason
&
Mark the target entry as deleted without removing its
history.
&
The target exists, and its deletion is justified by the
current state.
\\

Retrieve: query
&
Introduce relevant active entries from \(M_t\) into \(C_t\).
&
Deleted entries are excluded. The operation follows the configured
active-context budget.
\\

Filter: context items to remove
&
Remove the selected low-value items from \(C_t\).
&
The selected items belong to \(C_t\), and indispensable
evidence is preserved.
\\

SelectEpisode: historical fragments to restore
&
Introduce selected same-task fragments from \(H_t\) into
\(C_t\).
&
The fragments belong to the current task history and fit
the active-context budget.
\\

Summarize: context items and their summary
&
Replace selected content in \(C_t\) with a shorter summary.
&
The summary is faithful to its sources and preserves
task-relevant information.
\\

\bottomrule
\end{tabular}
\caption{
Compact specifications of the atomic memory tools in
VerMem. The table reports the information required by each
operation, its state transition, and its principal validity
condition.
}
\label{tab:app_atomic_tools}
\end{table*}

\paragraph{Add.}
Add stores new information with persistent utility. The new
entry records the information and its source. The operation
is rejected when the content is empty, unsupported, or
duplicates an existing active entry. When an existing entry
requires revision, Update is used instead.

\paragraph{Update.}
Update revises an existing LTM entry. It creates a new
version rather than destructively overwriting the previous
content. This retains the relation between the earlier and
current versions and allows later inspection of how the
memory state changed.

\paragraph{Delete.}
Delete follows soft-deletion semantics. The target entry is
excluded from subsequent ordinary retrieval, while its
content and deletion record remain available for inspection.
The retained record permits later inspection if the entry must be
reconsidered.

\paragraph{Retrieve.}
Retrieve is categorized as an STM tool because it does not
change \(M_t\). Instead, it selects active LTM entries and
introduces them into \(C_t\). The retrieval query is generated as part
of \(\xi_t\). Returned content is committed only when it satisfies the
configured active-context budget. Deleted entries cannot be retrieved.

\paragraph{Filter.}
Filter removes selected low-value content from \(C_t\).
Task instructions, the current environment state, and
indispensable evidence are protected. Filtering changes the
active context but does not remove the original event from
\(H_t\).

\paragraph{SelectEpisode.}
SelectEpisode performs episode-level context selection over
the current task history. It identifies earlier observations,
tool outputs, or intermediate conclusions that are relevant
to the current task state and returns them to \(C_t\). The
operation cannot access history from another task instance.

\paragraph{Summarize.}
Summarize replaces selected content in \(C_t\) with a
shorter representation. The summary must remain faithful to
the source content and preserve information required by the
current task. The original events remain available in
\(H_t\).

\subsection{Historical-Context Recovery via Episode-Level
Context Selection}
\label{app:historical_recovery}

Historical-context recovery concerns information that has
left the active context but becomes useful again at a later
task stage. It operates over the episodic history of the
current task. It is therefore distinct from retrieving
persistent information from LTM.

The system first identifies same-task historical fragments
that are relevant to the current task specification and active
context. The memory operation policy then selects the
fragments to be restored. Selected content is restored to
\(C_t\) subject to the configured active-context budget and retains its
original source and temporal position in \(H_t\).

The two information paths are
\[
M_t
\xrightarrow{\mathrm{Retrieve}}
C_t,
\qquad
H_t
\xrightarrow{\mathrm{SelectEpisode}}
C_t.
\]
Although both operations introduce information into the
active context, their information sources and intended uses
are different. Table~\ref{tab:app_retrieve_select}
summarizes this distinction.

\begin{table}[t]
\centering
\footnotesize
\setlength{\tabcolsep}{2.5pt}
\renewcommand{\arraystretch}{1.05}
\begin{tabular}{
@{}
p{0.19\columnwidth}
p{0.34\columnwidth}
p{0.37\columnwidth}
@{}}
\toprule
\textbf{Property}
&
\textbf{Retrieve}
&
\textbf{SelectEpisode}
\\
\midrule

Source
&
Long-term memory \(M_t\)
&
Episodic history \(H_t\)
\\

Content
&
Persistent facts and reusable knowledge
&
Earlier observations, tool outputs, and intermediate
conclusions
\\

Scope
&
The current LTM state
&
The current task history
\\

Motivation
&
Stored knowledge is absent from \(C_t\)
&
Earlier task evidence becomes relevant again
\\

Target
&
\(C_t\)
&
\(C_t\)
\\

Main risk
&
Irrelevant or outdated persistent information
&
Redundant or misleading historical fragments
\\

\bottomrule
\end{tabular}
\caption{
Difference between LTM retrieval and historical-context
recovery through episode-level context selection.
}
\label{tab:app_retrieve_select}
\end{table}

This separation supports same-task evidence recovery without treating
episodic history as another persistent memory store. It keeps Retrieve
and SelectEpisode as distinct operations.

\subsection{State Validation and Failure Handling}
\label{app:failure_handling}

The non-null memory operation is checked before it modifies
the real state. The checks cover the selected tool, required
arguments, referenced entries, operation preconditions,
source support, and available budget. The transition is
committed only when these conditions are satisfied.

If a proposed command is invalid, \(M_t\) and \(C_t\)
remain unchanged. The failed command and its reason are
retained in the task history. During training, the
corresponding structural, safety, or budget violation
contributes to \(c_{t,k}\). The task solver then continues
with the unchanged active context unless the task itself has
reached a terminal failure.

Update must preserve the relation between memory versions.
Delete must remain soft. Filter must not remove
indispensable evidence. SelectEpisode must remain within
the current task history. Summarize must reduce context
length without changing the supported meaning. Retrieve,
SelectEpisode, and Summarize must also satisfy the
active-context budget before their transitions are committed.

These execution checks are separate from semantic
verification. They determine whether an operation can be
executed safely. The local verifier instead evaluates the
task relevance, evidence grounding, local progress, and
information fidelity of a realized operation. The global verifier
supplies only evidence-coherence and terminal-memory-consistency
scores after trajectory completion. These scores enter the composite
global branch together with programmatic task, supporting-fact-recall,
and efficiency components. Both verifiers are disabled during inference.

\section{Optimization Objective and Loss Scope}
\label{app:optimization_loss_scope}

This section records the optimization settings and loss scope that
supplement the objectives in the main paper. Training-data and rollout
construction are detailed in Appendix~\ref{app:training_details};
verifier, normalization, and monitoring details are provided in
Appendix~\ref{app:verifier_credit_reward}.

\subsection{Core Training Configuration and Loss Scope}
\label{app:core_training_configuration}

Table~\ref{tab:app_training_configuration} summarizes the
settings directly related to data construction and trajectory
optimization.

\begin{table*}[t]
\centering
\small
\setlength{\tabcolsep}{5pt}
\renewcommand{\arraystretch}{1.05}
\begin{tabular}{
@{}
p{0.32\textwidth}
p{0.17\textwidth}
p{0.32\textwidth}
p{0.11\textwidth}
@{}}
\toprule
\textbf{Setting}
&
\textbf{Value}
&
\textbf{Setting}
&
\textbf{Value}
\\
\midrule

SFT train/dev size
&
32,000 / 4,000
&
Task batch per update
&
8
\\

SFT epochs
&
3
&
Candidate-group size \(K\)
&
8
\\

SFT learning rate
&
\(2\times10^{-5}\)
&
Phase A/B/C updates
&
1,000 / 1,000 / 1,500
\\

SFT effective batch
&
64
&
Rollout temperature
&
0.7
\\

SFT warmup ratio
&
0.03
&
Rollout top-\(p\)
&
0.95
\\

Maximum gradient norm
&
1.0
&
GRPO clip range
&
0.2
\\

RL learning rate
&
\(1\times10^{-6}\)
&
KL coefficient
&
0.1
\\

Policy-state serialization limit
&
8,192
&
Task-solver response limit
&
Configured protocol
\\

Memory-command output limit
&
Configured protocol
&
Retrieve/SelectEpisode/Summarize
&
Configured active-context budget
\\

Random seeds
&
42, 43, 44
&
Valid null accounting
&
Horizon: yes; tool calls: no
\\

\bottomrule
\end{tabular}
\caption{
Core training and rollout configuration used by VerMem.
Token limits are measured in model tokens.
}
\label{tab:app_training_configuration}
\end{table*}

During reinforcement learning, all generated tokens
belonging to one atomic memory operation share the same
\(A_{t,k}^{\mathrm{hier}}\). Different memory decisions
retain different advantages. The policy objective is applied
only to tokens that serialize the complete command \(u_{t,k}\),
including its structured arguments. Task-solver outputs,
environment observations,
input-state tokens, and padding tokens do not receive
memory-operation credit. Phase A, Phase B, and Phase C save separate checkpoints.
Each phase initializes the next one through policy parameters
only.

\section{Experimental Configuration and Evaluation Protocol}
\label{app:experimental_protocol}

The evaluation details below specify the splits, state
initialization, baseline configurations, metric computation,
Qwen-Max evaluation, budget-controlled analysis, and
multi-seed reporting used in the main experiments.
Training-data construction is described in
Appendix~\ref{app:training_details}, and the verifier and reward
implementations are described in
Appendix~\ref{app:verifier_credit_reward}. All methods are evaluated on the same task
instances with the same backbone, active-context budget,
and decoding configuration whenever backbone substitution
is supported.

\subsection{Benchmarks and Evaluation Splits}
\label{app:evaluation_splits}

We evaluate VerMem on ALFWorld, SciWorld, PDDL,
BabyAI, and HotpotQA. ALFWorld, SciWorld, PDDL, and
BabyAI use their official test splits. HotpotQA uses the
official distractor validation split as the held-out evaluation
set because it provides both reference answers and annotated
supporting facts. The HotpotQA training split is used only
for the SFT and reinforcement-learning instances described
in Appendix~\ref{app:training_details}. Original question identifiers do not overlap
between the construction-training pool, the internal
development pool, and the held-out evaluation set.

ALFWorld, SciWorld, and BabyAI use environment success
as the task outcome. PDDL uses the progress value returned
by the benchmark. One HotpotQA question defines one
evaluation episode. The question remains fixed as \(q\),
while the distractor passages enter the interaction as task
observations. Supporting-fact annotations are not exposed to the
memory operation policy. They are used for training-instance
construction, programmatic supporting-fact recall, verifier inputs,
and evaluation. The reference answer is likewise hidden from the
memory operation policy.

Table~\ref{tab:app_benchmark_protocol} summarizes the
evaluation split and reported metric for each benchmark.
Every executable instance in the selected split is evaluated.
All methods receive the same task order, environment
initialization, and information source.

\begin{table*}[t]
\centering
\small
\setlength{\tabcolsep}{5pt}
\renewcommand{\arraystretch}{1.05}
\begin{tabular}{
@{}
p{0.15\textwidth}
p{0.25\textwidth}
p{0.19\textwidth}
p{0.31\textwidth}
@{}}
\toprule
\textbf{Benchmark}
&
\textbf{Evaluation Split}
&
\textbf{Reported Metric}
&
\textbf{Evaluation Focus}
\\
\midrule

ALFWorld
&
Official test split
&
Success Rate
&
Long-horizon embodied interaction and environment-state
tracking.
\\

SciWorld
&
Official test split
&
Success Rate
&
Scientific procedure execution and maintenance of
intermediate observations.
\\

PDDL
&
Official AgentBoard test split
&
Progress Rate
&
Symbolic planning progress and intermediate goal-state
maintenance.
\\

BabyAI
&
Official test split
&
Success Rate
&
Instruction following, navigation, and short-term state
tracking.
\\

HotpotQA
&
Official distractor validation split
&
LLM-as-a-Judge \(J_{\mathrm{judge}}\)
&
Multi-hop evidence aggregation, persistent-memory
maintenance, and active-context control.
\\
\bottomrule
\end{tabular}
\caption{
Evaluation splits and primary metrics. Supporting-fact annotations are
not exposed to the memory operation policy. They are used for
training-instance construction, programmatic supporting-fact recall,
verifier inputs, and evaluation.
}
\label{tab:app_benchmark_protocol}
\end{table*}

\subsection{State Isolation and Episode Initialization}
\label{app:evaluation_state}

Every task instance uses an isolated memory state. Each
episode starts with empty \(M_0\) and \(H_0\), while \(C_0\)
contains only the task instruction and the initial input. The
initial input for ALFWorld, SciWorld, PDDL, and BabyAI is
the first observation returned by the benchmark. For
HotpotQA, it contains the question and the passage content
currently exposed by the task environment.

The memory states persist only within one episode. After
task completion, failure, or budget termination, \(M_t\),
\(C_t\), and \(H_t\) are released. The vector store, graph
state, metadata store, and temporary context of every
external baseline are also cleared before the next task
instance. Information from one evaluation example can
therefore never enter another example.

\subsection{Backbones and Inference Configuration}
\label{app:inference_configuration}

The main experiments use Qwen2.5-7B-Instruct and
Qwen3-4B-Instruct. VerMem, VerMem-noVerify, and Base
directly use the corresponding backbone. External methods
use the same backbone whenever their memory interface
supports backbone substitution. The corresponding method descriptions
and reported workflows are followed.

Evaluation uses deterministic decoding. The memory
operation policy and task solver use temperature \(0\) and
top-\(p\) \(1.0\). VerMem constructs \(x_t=g(s_t)\) under the
\(8{,}192\)-token policy-state serialization limit. The active-context
budget applies separately to \(C_t\). The configured protocol determines
the task-solver and memory-command output limits. Retrieve,
SelectEpisode, and Summarize follow the configured active-context
budget. The valid null command does not count as a memory-tool call
or in TC, but it remains a memory decision and counts toward the
memory-decision horizon.

Returned memory content is committed only when the resulting \(C_t\)
satisfies the configured active-context budget. Separately, VerMem uses
the bounded policy-state serialization in
Equation~\eqref{eq:state_serialization}; no additional ranking or
truncation rule is introduced here. The
local and global verifiers are disabled during
evaluation.

Table~\ref{tab:app_inference_settings} summarizes the
shared inference configuration.

\begin{table*}[t]
\centering
\small
\setlength{\tabcolsep}{5pt}
\renewcommand{\arraystretch}{1.05}
\begin{tabular}{
@{}
p{0.29\textwidth}
p{0.16\textwidth}
p{0.29\textwidth}
p{0.16\textwidth}
@{}}
\toprule
\textbf{Setting}
&
\textbf{Value}
&
\textbf{Setting}
&
\textbf{Value}
\\
\midrule

Policy temperature
&
0
&
Policy top-\(p\)
&
1.0
\\

Task-solver temperature
&
0
&
Task-solver top-\(p\)
&
1.0
\\

Policy-state serialization limit
&
8,192
&
Task-solver response limit
&
Configured protocol
\\

Memory-command output limit
&
Configured protocol
&
Retrieve/SelectEpisode/Summarize
&
Configured active-context budget
\\

Evaluation seeds
&
42, 43, 44
&
Local/global verifiers
&
Disabled
\\

Memory reset
&
After every episode
&
Null-decision accounting
&
Horizon: yes; TC: no
\\
\bottomrule
\end{tabular}
\caption{
Shared inference configuration. Token limits are measured
with the tokenizer of the corresponding backbone.
}
\label{tab:app_inference_settings}
\end{table*}

\subsection{Baseline Configurations}
\label{app:baseline_configurations}

Base uses the same task solver, backbone, response limit,
and active-context budget as VerMem, but it has no explicit
memory operation policy and no external persistent-memory
store.

For LangMem, A-Mem, Mem0, Mem0\(^{g}\), and AgeMem, the
corresponding method descriptions and reported workflows are followed.
LangMem retains modular
storage, search, and consolidation. A-Mem retains dynamic
indexing, linking, and memory evolution. Mem0 uses its
extraction, consolidation, and retrieval pipeline.
Mem0\(^{g}\) uses the graph-based memory workflow described for the
method. AgeMem retains its unified LTM/STM tool
policy. The final assembled prompt of every method is
subject to the shared evaluation context configuration. For VerMem,
the policy-state serialization limit and the active-context budget are
separate constraints.

VerMem-noVerify and VerMem share the architecture, backbone, complete
atomic command space, SFT initialization, Phase A/B/C curriculum, GRPO
configuration, and transition constraints. VerMem-noVerify uses the
normalized task-outcome advantage and hard constraints. It does not use
the local operation advantage or the composite global trajectory
advantage. VerMem uses the complete local--global credit assignment.

Table~\ref{tab:app_baseline_summary} summarizes the role
of each compared system. No method receives a reference
answer, supporting-fact label, or future observation during
inference.

\begin{table*}[t]
\centering
\footnotesize
\setlength{\tabcolsep}{4pt}
\renewcommand{\arraystretch}{1.05}
\begin{tabular}{
@{}
p{0.16\textwidth}
p{0.68\textwidth}
@{}}
\toprule
\textbf{Method}
&
\textbf{Evaluation Configuration}
\\
\midrule

Base
&
The common task solver without an explicit memory
operation policy or external persistent-memory store.
\\

LangMem
&
Modular memory workflow with storage, search, and consolidation.
\\

A-Mem
&
Dynamic indexing, linking, and memory-evolution workflow.
\\

Mem0
&
Extraction, consolidation, and retrieval pipeline.
\\

Mem0\(^{g}\)
&
Graph-based memory workflow described for Mem0\(^{g}\).
\\

AgeMem
&
Unified LTM/STM tool-based memory policy.
\\

VerMem-noVerify
&
The matched VerMem control that uses the normalized task-outcome
advantage and hard constraints, without the local operation advantage
or the composite global trajectory advantage.
\\

VerMem
&
The complete unified memory policy with local--global
verifier-guided credit assignment.
\\
\bottomrule
\end{tabular}
\caption{
Baseline and internal-control configurations. Every method
uses isolated task states and the same evaluation backbone
whenever backbone substitution is supported.
}
\label{tab:app_baseline_summary}
\end{table*}

\subsection{Metric Computation}
\label{app:metric_computation}

Success Rate on ALFWorld, SciWorld, and BabyAI is the
number of successful episodes divided by the number of evaluated
episodes. PDDL uses the normalized
Progress Rate returned by the benchmark and reports it on a
\(0\)--\(100\) scale. HotpotQA uses the Qwen-Max
LLM-as-a-Judge score \(J_{\mathrm{judge}}\), and the main table reports
\(100J_{\mathrm{judge}}\).

The Average column is the arithmetic mean of ALFWorld SR,
SciWorld SR, PDDL PR, BabyAI SR, and HotpotQA
\(100J_{\mathrm{judge}}\). Every benchmark therefore receives equal weight.

The reward-function ablation additionally reports TN, MQ,
and TC. TN is the mean number of online input and output
tokens from the memory operation policy and task solver in
one HotpotQA episode. MQ evaluates the active entries in
terminal \(M_t\) against the annotated supporting facts.
Soft-deleted entries are excluded. TC is the mean number of
non-null memory-tool calls. The null action
\(\varnothing\) is not counted in TC, but it remains part of the
memory-decision horizon.

\subsection{Qwen-Max Evaluation}
\label{app:qwen_evaluation}

The HotpotQA \(J_{\mathrm{judge}}\) and MQ metrics use the fixed
\texttt{qwen-max-2025-01-25} snapshot. The evaluator is
independent of the DeepSeek-V3.2 teacher and the local and
global verifiers. It does not generate SFT targets, provide
training rewards, or participate in a memory transition.

Qwen-Max follows the configured deterministic evaluation protocol. The
configured protocol determines its decoding and output limits. Every
example is evaluated independently, and the output must contain one
number in \([0,1]\). Malformed outputs follow the configured retry
handling.

The answer judge compares the question, ground-truth
answer, and agent answer. Semantically equivalent concise
answers can receive full credit. Partially correct answers
receive an intermediate score, while irrelevant or
contradictory answers receive a score close to zero.
The following block gives the complete answer-judge prompt. Bracketed
fields in this and subsequent prompt blocks denote values inserted at
runtime.

\paragraph{Qwen-Max answer-judge prompt.}
\begingroup\scriptsize
\begin{verbatim}
You are an independent judge evaluating the correctness
of an answer to a question.

Question:
[QUESTION]

Ground-truth answer:
[GROUND_TRUTH_ANSWER]

Agent answer:
[AGENT_ANSWER]

Score the agent answer from 0.0 to 1.0.

1.0: fully correct or semantically equivalent
0.8-0.9: correct with only a minor omission or wording issue
0.6-0.7: partially correct but missing an important element
0.4-0.5: contains some correct information and a major error
0.2-0.3: mostly incorrect with limited relevant content
0.0-0.1: incorrect, contradictory, or irrelevant

Return only one number between 0.0 and 1.0.
Do not provide an explanation.
\end{verbatim}
\endgroup

The MQ evaluator receives the question, reference answer,
annotated supporting facts, and active terminal LTM entries.
It evaluates supporting-fact coverage, relevance to the
question, and irrelevant stored content.
The following block gives the complete MQ prompt.

\paragraph{Qwen-Max memory-quality prompt.}
\begingroup\scriptsize
\begin{verbatim}
You are an independent judge evaluating the quality of
long-term memory for question answering.

Question:
[QUESTION]

Reference answer:
[REFERENCE_ANSWER]

Annotated supporting facts:
[SUPPORTING_FACTS]

Active entries in the terminal long-term memory:
[TERMINAL_LTM]

Evaluate the terminal long-term memory on three criteria:

1. Does it cover the supporting facts needed to answer the
   question?
2. Are the stored entries relevant and factually consistent?
3. Is it free of irrelevant, duplicated, or misleading
   content?

Score the memory from 0.0 to 1.0.

1.0: complete, relevant, consistent, and free of noise
0.8-0.9: nearly complete with only a minor omission or
         irrelevant entry
0.6-0.7: useful but missing important supporting content
0.4-0.5: contains some useful evidence and substantial noise
0.2-0.3: limited relevant content with major errors
0.0-0.1: unsupported, irrelevant, or unusable memory

Return only one number between 0.0 and 1.0.
Do not provide an explanation.
\end{verbatim}
\endgroup

MQ is used only for the reward-function analysis and is not
included in the Average column of the main results.

\subsection{Budget-Controlled Efficiency Evaluation}
\label{app:efficiency_protocol}

The efficiency experiment uses Qwen2.5-7B-Instruct and
compares A-Mem, Mem0, AgeMem, and VerMem on ALFWorld,
SciWorld, and BabyAI. Success Rate is computed separately
for the three benchmarks and then macro-averaged with equal
weight.

The online token budgets are
\(1{,}000\), \(1{,}500\), \(2{,}000\), \(2{,}500\),
\(3{,}000\), \(3{,}500\), \(4{,}000\), \(4{,}500\), and
\(5{,}000\). Each value is enforced as a hard episode
budget. The unfinished episode is counted as unsuccessful
when its budget is exhausted.

Online token accounting includes every input and output
token from the memory operation policy and task solver.
Repeated prompt content is counted again at every call.
Online generative memory modules used by an external
baseline are also included. Training, offline embedding,
index construction, local/global verifier calls, and
Qwen-Max evaluation are excluded.

We select \(S^*=40.0\) and \(B^*=2{,}500\) from the
validation curves and fix both values before test
evaluation. \(\mathrm{SR}@B^*\) is the macro-average SR at
the \(2{,}500\)-token budget. \(\mathrm{TN}@S^*\) is the linearly
interpolated online-token budget at which the mean success curve first
reaches \(S^*\), using the two adjacent evaluated budget points that
bracket the target.

Table~\ref{tab:app_efficiency_settings} summarizes the
fixed efficiency protocol. The lines in
Figure~\ref{fig:efficiency_frontier} show the mean across
three seeds, and the shaded regions denote one standard
deviation.

\begin{table}[t]
\centering
\small
\setlength{\tabcolsep}{5pt}
\renewcommand{\arraystretch}{1.05}
\begin{tabular}{@{}p{0.37\columnwidth}p{0.53\columnwidth}@{}}
\toprule
\textbf{Setting}
&
\textbf{Value}
\\
\midrule
Backbone
&
Qwen2.5-7B-Instruct
\\

Benchmarks
&
ALFWorld, SciWorld, BabyAI
\\

Methods
&
A-Mem, Mem0, AgeMem, VerMem
\\

Budget range
&
1,000--5,000
\\

Budget interval
&
500
\\

\(S^*\)
&
40.0
\\

\(B^*\)
&
2,500
\\

Budget termination
&
Unfinished episode is unsuccessful
\\

Curve aggregation
&
Mean and standard deviation over three seeds
\\
\bottomrule
\end{tabular}
\caption{
Fixed settings for the budget-controlled
efficiency--performance evaluation.
}
\label{tab:app_efficiency_settings}
\end{table}

\subsection{Random Seeds and Statistical Reporting}
\label{app:statistical_reporting}

Training and evaluation use seeds \(42\), \(43\), and
\(44\). Every seed independently completes SFT, Phase A,
Phase B, Phase C, and downstream evaluation. Methods use
the same benchmark instances and task order within one
seed.

The main result tables and quantitative appendix tables report the mean
across three seeds.
The shaded regions in Figures~\ref{fig:efficiency_frontier}
and~\ref{fig:reward_convergence} denote one standard
deviation.

Phase C checkpoints are selected only with the internal
development set. Official benchmark test results, held-out
HotpotQA \(J_{\mathrm{judge}}\), and MQ are never used for checkpoint
selection. Efficiency thresholds, evaluator prompts, and all
decoding settings are fixed before test evaluation.

\section{Training Data and Stateful Rollout Construction}
\label{app:training_details}

The training pipeline below expands the data construction, teacher
generation, quality filtering, and stateful trajectory
collection procedures introduced in the Training Pipeline.
We first describe how the HotpotQA training split is
converted into memory-decision states. We then specify the
SFT data composition, the frozen teacher prompt, and the
filtering protocol. The remaining subsections present phase-specific episode
construction and stateful candidate-trajectory collection.

\subsection{HotpotQA Training Data and Split}
\label{app:training_data_split}

All SFT and reinforcement learning instances are derived
from the HotpotQA training split~\citep{yang2018hotpotqa}.
Each original example provides a question \(q\), a reference
answer, annotated supporting facts, and distractor passages.
The question remains fixed within one training episode. The
reference answer supports instance construction and terminal task
signals, but it is not exposed to the memory operation policy.
Supporting-fact annotations are not exposed to the memory operation
policy. They are used for training-instance construction,
programmatic supporting-fact recall, verifier inputs, and evaluation.

We divide the original question identifiers in the HotpotQA
training split into a \(90\%\) construction-training pool and
a \(10\%\) internal-development pool. All states and episodes
derived from one original question remain in the same
partition. The official HotpotQA validation and test splits
are not used to construct SFT data or Phase A, B, or C
episodes.

The configured construction protocol converts supporting facts and
distractor passages into source-traceable evidence units used to validate
the atomic tool schemas. It also constructs the partial, conflicting,
duplicated, and obsolete training states needed by Update and Delete.
The exact representation fields and transformation rules are determined
by that protocol. These constructed states are not treated as native
HotpotQA annotations.

The final SFT training set contains \(32{,}000\) states, and
the internal development set contains \(4{,}000\) states.
Add, Update, Delete, Retrieve, Filter, SelectEpisode,
Summarize, and the null action \(\varnothing\) are balanced
in both sets. Each decision contributes \(4{,}000\) training
examples and \(500\) development examples. Half of each
class is derived from high-precision boundary templates, and
half is retained from teacher-generated candidates. At least
\(25\%\) of each class consists of boundary or negative
decision states.

\subsection{SFT Data Generation and Teacher Prompt}
\label{app:sft_teacher_generation}

Boundary templates construct memory-decision states with a
well-defined target. They emphasize distinctions that are
easy to confuse, including Add versus Update, Retrieve
versus SelectEpisode, Filter versus Summarize, and an
explicit operation versus \(\varnothing\). For example, when
a fact is already represented in \(M_t\) and no new evidence
is available, the target is \(\varnothing\) rather than a
duplicate Add. When the missing evidence appears only in
\(H_t\), the target is SelectEpisode rather than Retrieve.

The frozen DeepSeek-V3.2 teacher expands the diversity of
operation choices and structured arguments. For each state,
the teacher receives the bounded policy input
\(x_i\), the operation-type subspace allowed by the
current scenario, and the source
evidence already visible in the state. The reference answer
and future observations are not provided. Teacher-candidate sampling,
decoding, and output limits follow the configured protocol.

The following block gives the complete teacher prompt. The teacher
must return exactly one atomic
operation with its structured arguments, or
\(\varnothing\). It must not return an analysis, an
explanatory paragraph, or multiple candidate operations.

\paragraph{SFT teacher prompt.}
\begingroup\scriptsize
\begin{verbatim}
You are a teacher model for constructing supervised
examples for VerMem.

Given a task and the current memory-decision state, select
exactly one atomic memory operation, or select the null
action when no memory change is needed.

Use only information that is already visible in the state.
Do not use the reference answer, future observations, or
unsupported knowledge.

Available operations:
1. Add
   Use only when new information has persistent value and
   is not already represented in long-term memory.

2. Update
   Use when new evidence revises an existing long-term
   memory entry.

3. Delete
   Use only when an existing entry is clearly incorrect,
   obsolete, duplicated, or unsupported.

4. Retrieve
   Use when required information exists in long-term memory
   but is missing from the active context.

5. Filter
   Remove irrelevant active-context items while preserving
   all evidence needed by the current task.

6. SelectEpisode
   Restore task-relevant fragments from the episodic history
   of the current task.

7. Summarize
   Replace overlong active-context content with a shorter
   representation that preserves critical evidence.

8. Null action
   Use when the current memory state requires no change.

Output exactly one of the following forms:

Add(content="...", source_refs=[...])

Update(memory_id="...", new_content="...",
       source_refs=[...])

Delete(memory_id="...", reason="...")

Retrieve(query="...")

Filter(drop_refs=[...])

SelectEpisode(selected_refs=[...])

Summarize(source_refs=[...], summary="...")

\varnothing

Do not provide analysis, explanations,
markdown, or more than one operation.

Task:
[TASK]

Long-term memory:
[LONG_TERM_MEMORY]

Active context:
[ACTIVE_CONTEXT]

Episodic history:
[EPISODIC_HISTORY]

Allowed operation-type subspace:
[ALLOWED_ACTIONS]

Visible source evidence:
[VISIBLE_EVIDENCE]
\end{verbatim}
\endgroup

\subsection{SFT Quality Filtering and Optimization}
\label{app:sft_filtering}

Teacher outputs are not used directly as supervision labels.
Every candidate passes seven deterministic checks:
\begin{enumerate}
    \item the output contains exactly one operation;
    \item the operation belongs to the permitted operation-type
    subspace;
    \item all required arguments are present;
    \item every referenced memory entry, context item, or
    historical fragment exists;
    \item all references belong to the same HotpotQA
    training instance;
    \item the realized transition satisfies the tool
    semantics and preconditions in
    Appendix~\ref{app:method_tools};
    \item the output contains no answer leakage, future
    information, or unsupported content.
\end{enumerate}

Any candidate that fails a check is discarded without
automatic repair. Valid candidates are then deduplicated
according to their state, target operation, and structured
arguments. When two examples define equivalent states and
equivalent transitions, only one is retained.

Negative decision states never use an invalid operation as
the supervision target. The generator changes information
completeness, candidate location, distractor density, or
memory consistency so that a superficially plausible
operation becomes inappropriate. The target is another legal
operation, corrected arguments, or \(\varnothing\). A
duplicate Add may therefore be corrected to Update or
\(\varnothing\), while a Retrieve tendency is corrected to
SelectEpisode when the missing evidence appears only in
\(H_t\).

Table~\ref{tab:app_sft_coverage} reports the final SFT
coverage. Each target contributes \(4{,}000\) training
examples and \(500\) development examples.

\begin{table}[t]
\centering
\small
\setlength{\tabcolsep}{5pt}
\renewcommand{\arraystretch}{1.05}
\begin{tabular}{lrr}
\toprule
\textbf{Target Decision}
&
\textbf{Training}
&
\textbf{Development}
\\
\midrule
Add             & 4,000 & 500 \\
Update          & 4,000 & 500 \\
Delete          & 4,000 & 500 \\
Retrieve        & 4,000 & 500 \\
Filter          & 4,000 & 500 \\
SelectEpisode   & 4,000 & 500 \\
Summarize       & 4,000 & 500 \\
\(\varnothing\) & 4,000 & 500 \\
\midrule
\textbf{Total}  & \textbf{32,000} & \textbf{4,000} \\
\bottomrule
\end{tabular}
\caption{
Balanced target-command coverage in the SFT training and internal
development sets.
}
\label{tab:app_sft_coverage}
\end{table}

SFT uses AdamW for three epochs with a learning rate of
\(2\times10^{-5}\), a warmup ratio of \(0.03\), a maximum
gradient norm of \(1.0\), and an effective batch size of
\(64\). The policy-state serialization limit is \(8{,}192\) tokens. The
configured protocol determines the target-operation output limit. The loss is
applied only to the tokens that serialize the complete target
command \(u_i^\star=(v_i^\star,\xi_i^\star)\). Task descriptions,
input states, system
instructions, and padding tokens are excluded from the
target. Checkpoint selection uses tool accuracy and
executable-call rate on the development set. When two
checkpoints have the same tool accuracy, the one with the
lower invalid-operation rate is selected.

\subsection{Phase-Specific Episode Construction}
\label{app:phase_episode_construction}

Reinforcement learning follows the parameter curriculum in
Equation~\eqref{eq:parameter_curriculum}.
Each phase initializes its task states independently and
transfers only the policy parameters from the previous
phase. Table~\ref{tab:app_phase_construction} summarizes
the data size, operation-type subspace, and episode construction of
each phase.

\begin{table*}[t]
\centering
\footnotesize
\setlength{\tabcolsep}{3.5pt}
\renewcommand{\arraystretch}{1.06}
\begin{tabular}{
@{}
p{0.08\textwidth}
p{0.12\textwidth}
p{0.19\textwidth}
p{0.52\textwidth}
@{}}
\toprule
\textbf{Phase}
&
\textbf{Train/Dev}
&
\textbf{Operation-Type Subspace}
&
\textbf{Episode Construction}
\\
\midrule

Phase A
&
8,000 / 800
&
\(\mathcal V_{\mathrm{LTM}}
\cup\{\varnothing\}\)
&
Incremental facts, conflicting evidence, corrections,
duplicates, obsolete entries, and a delayed query.
\\

Phase B
&
8,000 / 800
&
\(\mathcal V_{\mathrm{STM}}
\cup\{\varnothing\}\)
&
Prepared LTM, distractor-rich active context, same-task
episodic history, summarization, and historical-context
recovery.
\\

Phase C
&
12,000 / 1,200
&
\(\mathcal V\cup\{\varnothing\}\)
&
Complete tasks with alternating LTM and STM operations;
some tasks additionally require historical-context recovery.
\\

\bottomrule
\end{tabular}
\caption{
Phase-specific episode construction. Only policy parameters
are transferred between phases; memory states are
initialized independently.
}
\label{tab:app_phase_construction}
\end{table*}

\paragraph{Phase A.}
Phase A contains \(8{,}000\) training episodes and \(800\)
development episodes. The initial \(M_0\) contains the
phase-specific persistent state, \(C_0\) contains the current
observation, and \(H_0\) begins with the current episode.
Add, Update, Delete, and \(\varnothing\) scenario families
are sampled uniformly. Supporting facts are revealed incrementally,
and each episode ends with a delayed
query. The rollout terminates when the delayed query is
completed, the token budget is exhausted, or another configured
termination condition is reached.

\paragraph{Phase B.}
Phase B contains \(8{,}000\) training episodes and \(800\)
development episodes. Each instance uses supporting documents and
distractor documents from the HotpotQA distractor setting. The initial
\(M_0\) contains phase-specific entries. The initial \(C_0\) occupies
part of the configured active-context budget, and \(H_0\) contains
same-task history. Retrieve, Filter, SelectEpisode,
Summarize, and \(\varnothing\) scenario families are sampled
uniformly.

Historical-context recovery cases place required evidence in earlier
same-task history while recent episodes contain related distractors.
Retrieve accesses \(M_t\), whereas SelectEpisode accesses \(H_t\).
Returned content must satisfy the configured active-context budget.
Valid null and rejected commands count toward the
memory-decision horizon; valid null commands do not count as
memory-tool calls.

\paragraph{Phase C.}
Phase C contains \(12{,}000\) training episodes and
\(1{,}200\) development episodes. Every episode requires at
least one LTM operation and one STM operation.
Some episodes require historical-context recovery so that
SelectEpisode receives supervision in complete tasks. Documents, task
feedback, and observations are revealed sequentially. Early Add or
Update decisions may affect a later Retrieve, while Filter or
Summarize may change the evidence available to subsequent
reasoning. Valid null and rejected commands count toward the
memory-decision horizon.

All phases use the \(8{,}192\)-token bounded policy-state serialization
defined by Equation~\eqref{eq:state_serialization}. The active-context
budget applies separately to \(C_t\). The configured protocol determines
the task-solver and memory-command output limits.

\subsection{Stateful Candidate-Trajectory Collection}
\label{app:stateful_collection}

At update \(n\) of phase \(p\in\{A,B,C\}\), the current
memory operation policy samples eight
tasks. For task \(q\), the generator constructs the
phase-specific initial state \(s_{0,p}^{(q)}\) and copies it
into \(K=8\) isolated rollouts according to
Equation~\eqref{eq:stateful_rollout}.
All trajectories in one candidate group share the task,
initial \(M_0\), \(C_0\), \(H_0\), and environment
condition. They differ only in policy sampling and the
resulting operation sequence. Rollouts use temperature
\(0.7\) and top-\(p\) \(0.95\).

At every step, the memory operation policy generates one
complete command \(u_{t,k}\), including \(\varnothing\). The valid
non-null command commits its transition through \(\mathcal T_m\). The
null command leaves \(M_t\) and \(C_t\) unchanged. On these states,
\(\mathcal T_m\) is the identity transition. The identity decision is
retained in the rollout record and does not create a memory-state
modification. If a
non-null command violates a precondition in
Appendix~\ref{app:method_tools}, \(M_t\) and \(C_t\) remain
unchanged, the failure record is added to \(H_t\), and the
corresponding \(c_{t,k}\) is recorded. The task solver then continues
with the resulting active context. Each rollout record retains the
bounded policy input \(x_{t,k}\), the complete command \(u_{t,k}\),
the post-memory state \(\widetilde s_t\), the post-interaction
state \(s_{t+1}\), and the command-token log probabilities. This
preserves the transition distinction made in
Equations~\eqref{eq:memory_command_transition}
and~\eqref{eq:environment_transition} without introducing a second
transition-record symbol.

Each trajectory terminates when the task
solver produces a final answer, the environment terminates,
the configured rollout limit is reached, or the
token budget is exhausted. Budget-terminated trajectories remain in the
candidate group and are processed by the composite global branch and
constraint channel. The global verifier contributes only its
evidence-coherence and terminal-memory-consistency components.

\paragraph{Stateful collection procedure.}
For every task in the current update, the collector copies
\(s_{0,p}^{(q)}\) into \(K\) isolated rollouts. Each rollout repeatedly
samples \(u_{t,k}\) from the current policy conditioned on
\(x_{t,k}\),
realizes or rejects \(\mathcal T_m\), applies the separate task and
environment transition, and stores both \(\widetilde s_t\) and
\(s_{t+1}\). The completed trajectories form
\(\mathcal G_{q,p}^{(n)}\) for local and composite-global credit
assignment.

Phase A and Phase B each use \(1{,}000\) policy updates,
while Phase C uses \(1{,}500\) updates. With eight tasks
and eight trajectories per task, the three phases collect
approximately \(64{,}000\), \(64{,}000\), and \(96{,}000\)
candidate trajectories, respectively.

\section{Additional Quantitative Analyses}
\label{app:additional_quantitative_results}

The analyses below provide detailed comparisons for the
main results, component study, verifier ablation, efficiency
evaluation, and reward-function ablation. Every value is
reported in the main paper or computed directly from the
reported results. No additional evaluation setting is
introduced.

\subsection{Detailed Cross-Backbone Comparison}
\label{app:cross_backbone_comparison}

Table~\ref{tab:app_cross_backbone_gains} compares VerMem
with the strongest external baseline for each benchmark and
backbone. With Qwen2.5-7B-Instruct, the absolute gains
range from \(3.67\) points on PDDL to \(8.31\) points on
HotpotQA. With Qwen3-4B-Instruct, the corresponding range
is \(3.68\) to \(8.12\) points. The average gains remain
close across the two backbones, reaching \(6.05\) and
\(5.54\) points. HotpotQA and SciWorld show the largest
improvements in both settings. Both benchmarks involve evidence
preservation, intermediate-state maintenance, and active-context
control. PDDL shows the
smallest gain, but VerMem remains above the strongest
external planning baseline under both backbones.

\begin{table*}[t]
\centering
\footnotesize
\setlength{\tabcolsep}{4.5pt}
\renewcommand{\arraystretch}{1.05}
\begin{tabular}{
@{}
l
l
c
l
c
c
@{}}
\toprule
\textbf{Backbone}
&
\textbf{Benchmark}
&
\textbf{VerMem}
&
\textbf{Strongest External Baseline}
&
\textbf{Baseline Score}
&
\textbf{Gain}
\\
\midrule

\multirow{6}{*}{Qwen2.5-7B-Instruct}
& ALFWorld & 46.30 & AgeMem & 41.07 & +5.23 \\
& SciWorld & 43.39 & AgeMem & 35.55 & +7.84 \\
& PDDL & 22.06 & A-Mem & 18.39 & +3.67 \\
& BabyAI & 65.55 & AgeMem & 61.42 & +4.13 \\
& HotpotQA & 62.75 & AgeMem & 54.44 & +8.31 \\
& Average & 48.01 & AgeMem & 41.96 & +6.05 \\
\midrule

\multirow{6}{*}{Qwen3-4B-Instruct}
& ALFWorld & 53.73 & AgeMem & 48.97 & +4.76 \\
& SciWorld & 66.37 & AgeMem & 59.48 & +6.89 \\
& PDDL & 38.75 & AgeMem & 35.07 & +3.68 \\
& BabyAI & 76.77 & AgeMem & 72.56 & +4.21 \\
& HotpotQA & 63.61 & AgeMem & 55.49 & +8.12 \\
& Average & 59.85 & AgeMem & 54.31 & +5.54 \\
\bottomrule
\end{tabular}
\caption{
Detailed comparison between VerMem and the strongest
external baseline for each benchmark and backbone. Gains
are absolute differences on the reporting scale used in the
main table.
}
\label{tab:app_cross_backbone_gains}
\end{table*}

\subsection{Effect of Local and Composite-Global Credit Across All Benchmarks}
\label{app:verifier_effect_all_tasks}

The main credit-branch ablation focuses on ALFWorld, SciWorld,
and HotpotQA. Table~\ref{tab:app_verify_gain_all_tasks}
compares full VerMem with the matched VerMem-noVerify control across
all five benchmarks. VerMem-noVerify uses the normalized task-outcome
advantage and hard constraints. It does not use the local operation
advantage or the composite global trajectory advantage. Adding both
credit branches raises four of the five task scores under
Qwen2.5-7B-Instruct and raises the Average by \(6.25\)
points. PDDL is the only exception, with a difference of
\(-0.35\) points. With Qwen3-4B-Instruct, VerMem improves
every task and raises the Average by \(6.21\) points. The
largest gains occur on HotpotQA and SciWorld for both
backbones. This pattern is consistent with complementary
operation-level and trajectory-level credit.

\begin{table*}[t]
\centering
\small
\setlength{\tabcolsep}{6pt}
\renewcommand{\arraystretch}{1.05}
\begin{tabular}{
@{}
l
cccccc
@{}}
\toprule
\textbf{Backbone}
&
\textbf{ALFWorld}
&
\textbf{SciWorld}
&
\textbf{PDDL}
&
\textbf{BabyAI}
&
\textbf{HotpotQA \(100J_{\mathrm{judge}}\)}
&
\textbf{Average}
\\
\midrule
Qwen2.5-7B-Instruct
& +6.12
& +9.45
& -0.35
& +6.34
& +9.67
& +6.25
\\
Qwen3-4B-Instruct
& +5.10
& +7.45
& +3.99
& +4.66
& +9.82
& +6.21
\\
\bottomrule
\end{tabular}
\caption{
Absolute difference between VerMem and VerMem-noVerify
across all tasks. Positive values favor VerMem.
VerMem-noVerify uses the normalized task-outcome advantage and hard
constraints; it does not use the local operation advantage or the
composite global trajectory advantage.
}
\label{tab:app_verify_gain_all_tasks}
\end{table*}

\subsection{Progressive Component Contributions}
\label{app:component_decomposition}

Table~\ref{tab:app_component_decomposition} reports the
score and marginal gain at every step of the cumulative
component study. The transition from Base to +LT produces
the largest initial improvement, especially on SciWorld.
Enabling the complete LTM/STM tool space yields further
gains on all three tasks. Stateful RL with the normalized task-outcome
advantage and hard constraints produces a larger gain on HotpotQA than
on the two interactive environments. The final transition from
+noV to +V adds the local operation advantage and the composite global
trajectory advantage. It improves
ALFWorld, SciWorld, and HotpotQA by \(6.12\), \(9.45\),
and \(9.67\) points. The progression is consistent with the two credit
branches building on unified memory control and stateful training.

\begin{table*}[t]
\centering
\footnotesize
\setlength{\tabcolsep}{4pt}
\renewcommand{\arraystretch}{1.05}
\begin{tabular}{
@{}
l
cc
cc
cc
@{}}
\toprule
\textbf{Configuration}
&
\textbf{ALFWorld}
&
\textbf{Increment}
&
\textbf{SciWorld}
&
\textbf{Increment}
&
\textbf{HotpotQA \(100J_{\mathrm{judge}}\)}
&
\textbf{Increment}
\\
\midrule
Base
& 27.16 & --
& 13.80 & --
& 38.36 & --
\\
+LT
& 37.73 & +10.57
& 27.98 & +14.18
& 45.76 & +7.40
\\
+LT/ST
& 39.02 & +1.29
& 31.26 & +3.28
& 48.73 & +2.97
\\
+noV
& 40.18 & +1.16
& 33.94 & +2.68
& 53.08 & +4.35
\\
+V
& 46.30 & +6.12
& 43.39 & +9.45
& 62.75 & +9.67
\\
\bottomrule
\end{tabular}
\caption{
Numerical decomposition of the cumulative component study.
Each increment is measured relative to the preceding
configuration. +noV uses the normalized task-outcome advantage and hard
constraints. +V adds the local operation advantage and the composite
global trajectory advantage and denotes full VerMem.
}
\label{tab:app_component_decomposition}
\end{table*}

\subsection{Complementarity of Local and Composite-Global Credit}
\label{app:verifier_complementarity}

Table~\ref{tab:app_verifier_complementarity} separates the
standalone and conditional gains of the two credit branches.
The local branch provides the larger standalone gain on
ALFWorld and SciWorld, while the composite global branch provides the
larger standalone gain on HotpotQA. The conditional differences are
consistent with complementary contributions after either branch has
been added. Adding composite-global credit after local credit yields
further gains of \(1.58\), \(3.58\), and \(4.49\) points.
Adding local credit after composite-global credit yields
\(2.99\), \(5.43\), and \(2.57\) points. The gains support
complementary roles for the two credit branches.

\begin{table*}[t]
\centering
\small
\setlength{\tabcolsep}{8pt}
\renewcommand{\arraystretch}{1.05}
\begin{tabular}{
@{}
p{0.37\textwidth}
ccc
@{}}
\toprule
\textbf{Comparison}
&
\textbf{ALFWorld}
&
\textbf{SciWorld}
&
\textbf{HotpotQA \(100J_{\mathrm{judge}}\)}
\\
\midrule
+Local over VerMem-noVerify
& +4.54 & +5.87 & +5.18
\\
+Global over VerMem-noVerify
& +3.13 & +4.02 & +7.10
\\
+Global after +Local
& +1.58 & +3.58 & +4.49
\\
+Local after +Global
& +2.99 & +5.43 & +2.57
\\
\bottomrule
\end{tabular}
\caption{
Standalone and conditional gains of the local and composite-global
credit branches under Qwen2.5-7B-Instruct. The composite global branch
contains programmatic and global-verifier components. +Local adds the
local operation advantage. +Global replaces the task-only trajectory
advantage with the composite global trajectory advantage.
}
\label{tab:app_verifier_complementarity}
\end{table*}

\subsection{Pairwise Efficiency Comparison}
\label{app:pairwise_efficiency}

Table~\ref{tab:app_pairwise_efficiency} expands the two
efficiency metrics into pairwise comparisons. VerMem
reduces the token cost required to reach \(S^*=40.0\) by
\(20.0\%\) relative to AgeMem, \(45.5\%\) relative to
Mem0, and \(51.1\%\) relative to A-Mem. Under the fixed
budget \(B^*=2{,}500\), it improves macro-average SR by
\(5.30\), \(8.70\), and \(10.50\) points. The ordering is
consistent under both metrics and supports a stronger
efficiency--performance frontier for VerMem.

\begin{table}[t]
\centering
\footnotesize
\setlength{\tabcolsep}{4pt}
\renewcommand{\arraystretch}{1.05}
\begin{tabular}{
@{}
l
cccc
@{}}
\toprule
\textbf{Method}
&
\(\mathbf{\mathrm{TN}@S^*}\)
&
\textbf{Reduction}
&
\(\mathbf{\mathrm{SR}@B^*}\)
&
\textbf{VerMem Gain}
\\
\midrule
A-Mem
& 4,250 & 51.1\% & 33.80 & +10.50
\\
Mem0
& 3,820 & 45.5\% & 35.60 & +8.70
\\
AgeMem
& 2,600 & 20.0\% & 39.00 & +5.30
\\
VerMem
& 2,080 & -- & 44.30 & --
\\
\bottomrule
\end{tabular}
\caption{
Pairwise efficiency comparison under
\(S^*=40.0\) and \(B^*=2{,}500\). Reduction reports the
relative decrease in \(\mathrm{TN}@S^*\) achieved by VerMem.
}
\label{tab:app_pairwise_efficiency}
\end{table}

\subsection{Reward-Function Effects}
\label{app:reward_effects}

Table~\ref{tab:app_reward_changes} reports the absolute and
relative changes associated with All-Returns. The complete
reward raises \(J_{\mathrm{judge}}\) by \(0.097\) and MQ by \(0.182\), which
correspond to relative improvements of \(18.3\%\) and
\(37.1\%\). TN decreases by \(168\) tokens, or \(7.14\%\),
while TC increases by \(1.16\) calls. The simultaneous
increase in TC and reduction in TN is inconsistent with uniform
tool-call suppression as the sole explanation. These aggregate metrics
do not identify operation-specific contributions.

\begin{table*}[t]
\centering
\small
\setlength{\tabcolsep}{10pt}
\renewcommand{\arraystretch}{1.05}
\begin{tabular}{
@{}
l
cccc
@{}}
\toprule
\textbf{Metric}
&
\textbf{Answer-Only}
&
\textbf{All-Returns}
&
\textbf{Absolute Change}
&
\textbf{Relative Change}
\\
\midrule
\(J_{\mathrm{judge}}\uparrow\)
& 0.531 & 0.628 & +0.097 & +18.3\% \\
TN\(\downarrow\)
& 2,352 & 2,184 & -168 & -7.14\% \\
MQ\(\uparrow\)
& 0.491 & 0.673 & +0.182 & +37.1\% \\
TC
& 4.47 & 5.63 & +1.16 & +26.0\% \\
\bottomrule
\end{tabular}
\caption{
Absolute and relative effects of the complete reward on
HotpotQA using Qwen2.5-7B-Instruct. Negative TN change is
favorable.
}
\label{tab:app_reward_changes}
\end{table*}

The reported ablations associate the complete command space,
task-outcome stateful RL, and the complete training signal with higher
scores across both evaluated backbones. VerMem also attains the stronger
reported efficiency frontier. The results support the joint role of
unified memory management and multi-granularity credit assignment.

\section{Verifier, Credit, Reward, and Reproducibility Specifications}
\label{app:verifier_credit_reward}

The verifier specification below covers the local and global verifiers,
operation-wise normalization, the constraint channel, and the
command-token loss mask and monitoring protocol introduced in the main
paper. The local verifier evaluates one realized atomic
memory transition. The global verifier supplies evidence-coherence and
terminal-memory-consistency scores after task completion; it does not
compute task correctness, supporting-fact recall, or efficiency.
Structural validity, state-transition
preconditions, safety requirements, and budget limits are
handled by the constraint channel rather than the semantic
verifier scores. This separation keeps semantic and hard-constraint
penalties distinct.

\subsection{Verifier Instantiation and Scoring Protocol}
\label{app:verifier_protocol}

The local and global verifiers use frozen DeepSeek-V3.2
instances with separate prompts and input scopes. The local
verifier receives the task \(q\), the bounded policy-visible state
\(x_t=g(s_t)\), the complete command \(u_t\), the realized memory
transition, and the source
evidence visible at the current decision. It does not receive an
unbounded concatenation of \(H_t\), nor does it observe later memory
decisions, the final answer, or the global-verifier output.
The global verifier is invoked after trajectory termination.
Its input contains the task, reference answer, annotated
supporting facts, final answer, ordered memory commands,
their realized transitions, the terminal \(M_t\), \(C_t\),
and \(H_t\). Online cost is computed programmatically and is not
provided as a verifier-produced score.

Verifier-input truncation, decoding, and output limits are determined by
the configured protocol. No separate ranking, removal priority, or
compression rule is specified here.

Table~\ref{tab:app_verifier_configuration} summarizes the
verifier configuration.

\begin{table*}[t]
\centering
\small
\setlength{\tabcolsep}{5pt}
\renewcommand{\arraystretch}{1.05}
\begin{tabular}{
@{}
p{0.28\textwidth}
p{0.27\textwidth}
p{0.27\textwidth}
@{}}
\toprule
\textbf{Setting}
&
\textbf{Local Verifier}
&
\textbf{Global Verifier}
\\
\midrule
Model
&
Frozen DeepSeek-V3.2
&
Frozen DeepSeek-V3.2
\\

Maximum output
&
Configured protocol
&
Configured protocol
\\

Decoding
&
Configured protocol
&
Configured protocol
\\

Output dimensions
&
Relevance, grounding, local progress, and information
fidelity
&
Evidence coherence and terminal-memory consistency
\\

Score range
&
Integer \(0\)--\(4\)
&
Integer \(0\)--\(4\)
\\

Format retry
&
One deterministic retry
&
One deterministic retry
\\
\bottomrule
\end{tabular}
\caption{
Configuration of the local and global verifiers.
The two verifiers use separate prompts and are disabled
during evaluation.
}
\label{tab:app_verifier_configuration}
\end{table*}

Each verifier dimension is scored with an integer in
\(\{0,1,2,3,4\}\). Score \(0\) denotes a contradicted
or harmful decision; score \(1\) denotes weak support
with major errors; score \(2\) denotes a partially
justified decision with clear omissions; score \(3\)
denotes a sound decision with minor deficiencies; and score
\(4\) denotes a fully justified decision. Scores are
divided by \(4\) before reward computation.

The implementation performs deterministic parsing and retry handling.
The exact fallback after repeated malformed verifier outputs is
determined by the execution protocol.

\subsection{Local Verifier}
\label{app:local_verifier}

The local verifier evaluates one executable complete command and its
realized state transition. Executable commands include valid null
decisions, which use the separate null rubric below. It scores task relevance,
evidence grounding, local progress, and information
fidelity. The four scores are divided by \(4\) and averaged
to obtain \(r_{t,k}^{\mathrm{local}}\).
The trajectory-level local reward in
Equation~\eqref{eq:local_reward} sums the scores produced for executable
decisions in \(\mathcal E_k\) and normalizes them by the full
memory-decision horizon \(T_k\). It is used only for monitoring. It does
not replace the operation-wise normalization in
Equation~\eqref{eq:normalized_advantages} used to compute
\(A_{t,k}^{\mathrm{local}}\).

Task relevance measures whether the command addresses an
actual memory need in the bounded state \(x_t\). Evidence grounding measures
whether its content, target, and source are supported by the
currently visible evidence. Local progress measures whether
the transition improves the current task state. Information
fidelity measures whether the operation preserves valid
facts, relations, and source attribution.

Table~\ref{tab:app_local_rubric} gives the tool-specific
conditions for the highest local score. Lower scores reflect
increasing departures from these conditions.

\begin{table*}[t]
\centering
\footnotesize
\setlength{\tabcolsep}{4pt}
\renewcommand{\arraystretch}{1.05}
\begin{tabular}{
@{}
p{0.13\textwidth}
p{0.73\textwidth}
@{}}
\toprule
\textbf{Decision}
&
\textbf{Condition for a Fully Justified Transition}
\\
\midrule

Add
&
The content has persistent utility, follows visible source
evidence, and is not already represented by an active entry
in \(M_t\).
\\

Update
&
New evidence revises an existing entry, the target is
correct, and valid information not contradicted by the new
evidence is preserved.
\\

Delete
&
The target is incorrect, unsupported, duplicated, or
obsolete, and soft deletion does not remove information that
remains necessary for the task.
\\

Retrieve
&
The returned active LTM entries fill a current information
gap in \(C_t\) without introducing substantial irrelevant or
outdated content.
\\

Filter
&
The removed items are irrelevant or redundant, while the
task instruction, current environment state, and
indispensable evidence remain in \(C_t\).
\\

SelectEpisode
&
The selected fragments belong to the current task history
and restore earlier observations, tool outputs, or
intermediate conclusions required by the current task state.
\\

Summarize
&
The operation reduces active-context length while preserving
critical facts, relations, and source attribution.
\\

\(\varnothing\)
&
The current state requires no memory transition. The null command leaves
\(M_t\) and \(C_t\) unchanged. The identity decision is retained in the
rollout record and does not create a memory-state modification.
\\
\bottomrule
\end{tabular}
\caption{
Operation-specific rubric used by the local verifier.
Structural validity and hard constraints are evaluated
separately.
}
\label{tab:app_local_rubric}
\end{table*}

Commands rejected by the execution checks in
Appendix~\ref{app:method_tools}
are not sent to the local verifier. Their
\(A_{t,k}^{\mathrm{local}}\) is set to zero. They still receive the
trajectory-level \(A_k^{\mathrm{global}}\), while the corresponding
\(c_{t,k}\) supplies the direct decision-level penalty.

The following block gives the local verifier prompt. The
operation-specific condition is selected
from Table~\ref{tab:app_local_rubric}.

\paragraph{Local-verifier prompt.}
\begingroup\scriptsize
\begin{verbatim}
You are the local verifier used during VerMem training.

Evaluate exactly one realized atomic memory transition.
The transition has already passed the structural,
reference, precondition, and budget checks.

Use only the task, the state before the command, the
selected command, the realized state after the command,
and the visible source evidence.

Do not use a later trajectory state, the final answer, or
information that is absent from the supplied state.

Score four dimensions with integers from 0 to 4.

0: contradicted, harmful, or entirely unjustified
1: weakly supported with major errors
2: partially justified with clear omissions
3: sound with minor deficiencies
4: fully justified for the current state

RELEVANCE
Does the command address the current memory need?

GROUNDING
Is the command supported by visible source evidence?

LOCAL_PROGRESS
Does the realized transition improve the current task or
memory state?

INFORMATION_FIDELITY
Does the transition preserve the valid information required
from its sources?

Output exactly four lines and no additional text:

RELEVANCE: <0-4>
GROUNDING: <0-4>
LOCAL_PROGRESS: <0-4>
INFORMATION_FIDELITY: <0-4>

Task q:
[TASK]

Bounded state before the command x_t:
[STATE_BEFORE]

Complete memory command u_t:
[OPERATION]

State after the command:
[STATE_AFTER]

Visible source evidence:
[VISIBLE_EVIDENCE]

Operation-specific condition:
[OPERATION_CONDITION]
\end{verbatim}
\endgroup

\subsection{Composite Global Branch and Global Verifier}
\label{app:global_verifier}

The composite global branch follows
Equations~\eqref{eq:global_reward}
and~\eqref{eq:evidence_state_rewards}. It combines task correctness,
supporting-fact recall, evidence coherence, terminal-memory
consistency, and trajectory efficiency. The global verifier produces
only evidence coherence and terminal-memory consistency; the other
components are programmatic. The four resulting reward terms in
Equation~\eqref{eq:global_reward} are normalized to \([0,1]\) and
averaged with equal weight.

VerMem-noVerify uses the normalized task-outcome advantage and hard
constraints. It does not use the local operation advantage or the
composite global trajectory advantage.

For HotpotQA, \(r_k^{\mathrm{task}}\) is the mean of exact
match and token-level F1 against the reference answer. Both
values are normalized to \([0,1]\). If no final answer is
produced, \(r_k^{\mathrm{task}}\) is set to zero.

The evidence component uses both annotated supporting facts and global
verification. Supporting-fact annotations are not exposed to the memory
operation policy. They are used for training-instance construction,
programmatic supporting-fact recall, verifier inputs, and evaluation.
The exact evidence extraction procedure follows the implementation
protocol and is not redefined here. The supporting-fact recall is averaged with the normalized
evidence-coherence score returned by the global verifier to
obtain \(r_k^{\mathrm{evid}}\). Evidence coherence measures
whether the final answer follows from a complete,
task-relevant, and non-contradictory evidence chain.

The global verifier also returns terminal-memory
consistency. Its normalized score forms
\(r_k^{\mathrm{state}}\). This dimension checks whether the
terminal LTM, active context, and episodic history remain
consistent with the observed evidence, whether critical
supporting facts remain accessible, and whether obsolete or
conflicting entries have been handled. Because \(H_t\) is an ordered
provenance record, superseded, soft-deleted, rejected, or failed events
may remain in history when their status is explicit.

The normalized online cost \(\overline C_k^{\mathrm{online}}\) uses
online tokens, task steps, and memory-tool calls. The three
costs are independently min--max normalized within the
\(K=8\) candidate trajectories sampled for the same task and
then averaged. Each cost component is set to zero when all
candidate trajectories have the same value for that
component. Teacher calls, verifier calls, Qwen-Max
evaluation, and offline index construction are excluded.

The completion threshold is \(\delta=0.5\), as used in
Equation~\eqref{eq:efficiency_reward}.
The efficiency component is available only when
\(r_k^{\mathrm{task}}\geq\delta\). Under this gate, a low-cost failed
trajectory receives no positive efficiency reward.

The global verifier does not rescore tool syntax, required
arguments, or hard preconditions. Task completion,
supporting-fact recall, and online cost are computed
programmatically. The verifier evaluates only evidence
coherence and terminal-memory consistency.

The following block gives the global verifier prompt.

\paragraph{Global-verifier prompt.}
\begingroup\scriptsize
\begin{verbatim}
You are the global verifier used during VerMem training.

Evaluate the completed task trajectory as a whole.

Answer correctness, supporting-fact recall, online cost,
tool syntax, and hard preconditions are computed
separately. Do not rescore these items.

Score two dimensions with integers from 0 to 4.

0: contradicted, incoherent, or seriously inconsistent
1: major evidence or terminal-memory failures
2: partially coherent with important omissions
3: coherent with minor deficiencies
4: complete, well supported, and internally consistent

EVIDENCE_COHERENCE
Does the final answer follow from a complete,
task-relevant, and non-contradictory evidence chain?

TERMINAL_MEMORY_CONSISTENCY
Are the terminal long-term memory, active context, and
episodic history consistent with the observed evidence?
Are critical facts preserved and accessible? Are obsolete
or conflicting entries handled appropriately? Historical
superseded, deleted, rejected, or failed events may remain
when their status is explicit.

Output exactly two lines and no additional text:

EVIDENCE_COHERENCE: <0-4>
TERMINAL_MEMORY_CONSISTENCY: <0-4>

Task q:
[TASK]

Reference answer:
[REFERENCE_ANSWER]

Annotated supporting facts:
[SUPPORTING_FACTS]

Final answer:
[FINAL_ANSWER]

Ordered memory commands and realized transitions:
[MEMORY_TRAJECTORY]

Terminal long-term memory:
[TERMINAL_LTM]

Terminal active context:
[TERMINAL_CONTEXT]

Terminal episodic history:
[TERMINAL_HISTORY]
\end{verbatim}
\endgroup

\subsection{Advantage Normalization and Hierarchical Credit}
\label{app:advantage_normalization}

The composite global score is normalized within the \(K=8\)
trajectories sampled from the same task. Local scores are
normalized by operation type \(v_{t,k}\), as defined in
Equation~\eqref{eq:normalized_advantages}. The seven atomic tools and
the null type \(v=\varnothing\) maintain separate local statistics.

When an operation type appears at least eight times in the
current update, its current-update mean and standard
deviation are used. When fewer than eight instances are
available, phase-specific running statistics are used. Before
each phase, the running statistics are initialized from
\(128\) executable development transitions for every
decision type enabled in that phase. The statistics are
updated with an exponential moving-average decay of
\(0.99\). They are reinitialized at the start of every phase.

Both local and global normalization use a standard-deviation
floor of \(0.01\), and the \(\epsilon\) in the main paper is
set to \(10^{-6}\). If every trajectory in a candidate group
has the same composite global score, all corresponding global
advantages are zero. Local and global advantages are clipped
to \([-5,5]\) before they are combined.

The three coefficients retain the unit values specified with
Equation~\eqref{eq:hierarchical_credit}. Each rejected command receives
\(A_{t,k}^{\mathrm{local}}=0\). Its global advantage remains
available because the completed trajectory still determines
\(A_k^{\mathrm{global}}\). The
constraint cost is assigned directly to the corresponding
memory decision.

Table~\ref{tab:app_credit_settings} summarizes the fixed
normalization and optimization settings.

\begin{table*}[t]
\centering
\small
\setlength{\tabcolsep}{5pt}
\renewcommand{\arraystretch}{1.05}
\begin{tabular}{
@{}
p{0.31\textwidth}
p{0.14\textwidth}
p{0.31\textwidth}
p{0.14\textwidth}
@{}}
\toprule
\textbf{Setting}
&
\textbf{Value}
&
\textbf{Setting}
&
\textbf{Value}
\\
\midrule

Candidate-group size \(K\)
&
8
&
Minimum current-update count
&
8
\\

Running-statistic initialization
&
128 per decision
&
Running-statistic decay
&
0.99
\\

Standard-deviation floor
&
0.01
&
\(\epsilon\)
&
\(10^{-6}\)
\\

Advantage clipping
&
\([-5,5]\)
&
\(\delta\)
&
0.5
\\

\(\lambda_{\mathrm{local}}\)
&
1.0
&
\(\lambda_{\mathrm{global}}\)
&
1.0
\\

\(\lambda_{\mathrm{constraint}}\)
&
1.0
&
\(\epsilon_{\mathrm{clip}}\)
&
0.2
\\

\(\beta_{\mathrm{KL}}\)
&
0.1
&
Entropy coefficient
&
0
\\
\bottomrule
\end{tabular}
\caption{
Fixed normalization, credit-assignment, and stabilization
settings used by VerMem.
}
\label{tab:app_credit_settings}
\end{table*}

\subsection{Auxiliary Constraints}
\label{app:auxiliary_constraints}

The constraint channel handles discrete and deterministic
violations. Table~\ref{tab:app_constraint_costs} reports the
fixed costs and execution behavior. Multiple violations at
one memory decision are capped at
\(c_{t,k}=1.0\).
The valid null decision counts toward the memory-decision horizon used
by the auxiliary cost, but it is not a memory-tool call and does not
incur a tool-call cost.

\begin{table*}[t]
\centering
\footnotesize
\setlength{\tabcolsep}{4pt}
\renewcommand{\arraystretch}{1.05}
\begin{tabular}{
@{}
p{0.48\textwidth}
p{0.09\textwidth}
p{0.33\textwidth}
@{}}
\toprule
\textbf{Violation}
&
\(\mathbf{c_{t,k}}\)
&
\textbf{Execution Behavior}
\\
\midrule

Invalid tool, missing argument, invalid reference, or failed
precondition
&
1.0
&
Reject the transition, keep \(M_t\) and \(C_t\) unchanged,
record the failed command and reason in \(H_t\), and continue the task.
\\

Unsupported Add or Update, or an unjustified Delete
&
1.0
&
Reject the transition, keep \(M_t\) and \(C_t\) unchanged, and
record the failure in \(H_t\).
\\

Unrecoverable removal of indispensable evidence
&
1.0
&
Reject the proposed transition and preserve the previous \(M_t\) and
\(C_t\); record the failure in \(H_t\).
\\

Active-context overflow
&
0.5
&
Reject the transition, keep \(M_t\) and \(C_t\) unchanged, and record
the failed command and status in \(H_t\).
\\

Memory-decision or interaction-limit violation
&
1.0
&
Terminate the trajectory at the configured limit.
\\

Reference-answer or future-information leakage
&
1.0
&
Terminate the trajectory and set its task and evidence
components to zero.
\\
\bottomrule
\end{tabular}
\caption{
Constraint costs and execution behavior. Costs from
multiple violations at one memory decision are capped at
\(1.0\).
}
\label{tab:app_constraint_costs}
\end{table*}

Evidence grounding and information fidelity remain
continuous semantic scores. The unsupported-operation
constraint is triggered only when no valid source exists or
the operation directly contradicts visible evidence. The
evidence-loss constraint is triggered only when
indispensable information is removed and cannot be
recovered from \(M_t\) or \(H_t\). Partial omissions lower
the local score but do not trigger a hard constraint.

\subsection{Token-Level GRPO Implementation}
\label{app:token_grpo}

The policy ratio is evaluated token by token as
\(\rho_{t,k,j}(\theta)\) in
Equation~\eqref{eq:token_ratio_kl}. All generated tokens that serialize
one complete command \(u_{t,k}\) share the same
\(A_{t,k}^{\mathrm{hier}}\). The token loss of one command is
averaged over its command tokens before losses are
averaged over \(\mathcal B_{\mathrm{op}}\). This gives each command
equal weight before batch averaging, independent of argument length.

The loss mask includes only the complete command tokens
\(y_{t,k,1:L_{t,k}}\), including the operation type and structured
arguments. The task specification, \(M_t\), \(C_t\), \(H_t\),
system instructions, task-solver outputs, environment
observations, and padding tokens are excluded. The null
action \(\varnothing\) remains a trainable memory decision,
and it counts toward the memory-decision horizon. It is not counted as
a memory-tool call.

The old policy is frozen before each policy update. The
reference policy remains fixed within one phase. Phase A,
Phase B, and Phase C use
\(\theta_{\mathrm{SFT}}\), \(\theta_A\), and \(\theta_B\),
respectively, as their reference policies.

The objective uses \(\epsilon_{\mathrm{clip}}=0.2\) and
\(\beta_{\mathrm{KL}}=0.1\), as specified in
Equation~\eqref{eq:grpo_objective}.
The reinforcement-learning rate is \(10^{-6}\), the maximum
gradient norm is \(1.0\), and no additional entropy bonus is
used. Local and global advantages are clipped to
\([-5,5]\).

\subsection{Reward Monitoring and Convergence Analysis}
\label{app:reward_monitoring}

Figure~\ref{fig:reward_convergence} compares the
strategy-specific monitoring values of Answer-Only and All-Returns.
They use the mappings \(m_k^{\mathrm{ans}}\) and
\(m_k^{\mathrm{all}}\) in
Equations~\eqref{eq:answer_only_monitor}
and~\eqref{eq:all_returns_monitor}. Since the two strategies use
different monitoring definitions, their absolute values are not
required to
coincide at the first recorded step.

The complete reinforcement-learning curriculum contains
\(3{,}500\) policy updates. We record \(101\) training points.
Training Step \(0\) is the rollout batch collected before the
first policy update, and consecutive recorded points
are separated by \(35\) cumulative updates. At every
recorded point, the monitoring value is averaged over the \(64\) candidate
trajectories in the current rollout batch. The plotted line is
the mean across seeds \(42\), \(43\), and \(44\), and the
shaded region denotes one standard deviation across the
three runs.

For the common \([0,1]\) axis, Answer-Only plots
\(m_k^{\mathrm{ans}}\), while All-Returns plots
\(m_k^{\mathrm{all}}\), whose global term is
\(r_k^{\mathrm{global}}\). These mappings are used only for training
visualization. They are not policy advantages and do not change the
policy objective or checkpoint selection.

The curves use the raw recorded-point statistics without a
moving average, temporal smoothing, or spline
interpolation. White squares mark every tenth recorded
point. The observed curves are consistent with earlier stabilization
and lower late-stage variability for the All-Returns configuration,
but the monitoring mappings differ and their absolute levels are not
directly comparable.

\subsection{Protocol-Level Behavior and Failure Analysis}
\label{app:qualitative_analysis}

This section organizes protocol-level memory-decision patterns and
failure cases implied by the tool semantics and reported ablations. It
does not introduce another metric or claim an additional empirical case
study.

\subsubsection{Phase-Specific Memory Decisions}
\label{app:phase_specific_behavior}

The three training phases expose different memory needs.
Phase A presents incremental facts, corrections, duplicate
content, and obsolete entries. The correct decision therefore
depends on whether the visible evidence should create,
revise, remove, or leave unchanged an entry in \(M_t\).
Phase B presents a prepared LTM state, a distractor-rich
\(C_t\), and same-task history in \(H_t\). It requires the
policy to distinguish Retrieve from SelectEpisode and Filter
from Summarize. Phase C combines both tool groups in one
trajectory. Early Add or Update decisions may support a later
Retrieve, while Filter, Summarize, or SelectEpisode changes
the evidence visible to the task solver.

Table~\ref{tab:app_phase_decisions} summarizes the
decision patterns that define these scenarios. The state
effects follow the tool semantics in
Appendix~\ref{app:method_tools}.

\begin{table*}[t]
\centering
\footnotesize
\setlength{\tabcolsep}{4pt}
\renewcommand{\arraystretch}{1.05}
\begin{tabular}{
@{}
p{0.24\textwidth}
p{0.13\textwidth}
p{0.31\textwidth}
p{0.24\textwidth}
@{}}
\toprule
\textbf{Memory Need}
&
\textbf{Decision}
&
\textbf{Realized State Effect}
&
\textbf{Targeted Failure Mode}
\\
\midrule

New information has persistent value and is absent from
active LTM.
&
Add
&
Create a source-bearing entry in \(M_t\).
&
Loss of information required by a delayed query.
\\

New evidence corrects or refines an active LTM entry.
&
Update
&
Create a revised version while preserving the previous
version.
&
Stale or conflicting persistent memory.
\\

The active entry is unsupported, obsolete, or duplicated.
&
Delete
&
Soft-delete the target while retaining its history.
&
Repeated retrieval of invalid or redundant content.
\\

Required information is present in \(M_t\) but absent from
\(C_t\).
&
Retrieve
&
Introduce the relevant active LTM content into \(C_t\).
&
Reasoning without available persistent evidence.
\\

The active context contains irrelevant or redundant
distractors.
&
Filter
&
Remove selected low-value items from \(C_t\) while
preserving their history in \(H_t\).
&
Context pollution and unnecessary online tokens.
\\

Required same-task evidence exists only in earlier
\(H_t\).
&
SelectEpisode
&
Restore selected historical fragments to \(C_t\).
&
Failure of a recent-window or LTM-only recovery strategy.
\\

Relevant context is too long for efficient continued
reasoning.
&
Summarize
&
Replace selected content in \(C_t\) with a shorter faithful
representation.
&
Context overflow or repeated processing of the same
evidence.
\\

The current state has no valid memory need.
&
\(\varnothing\)
&
Leave \(M_t\) and \(C_t\) unchanged and create no memory-state
modification. Retain the identity decision in the rollout record. It
counts toward the memory-decision horizon, but not toward memory-tool
calls or TC.
&
Unnecessary operations and additional tool cost.
\\
\bottomrule
\end{tabular}
\caption{
Qualitative decision patterns covered by the three-stage
training curriculum. Each row links a memory need to the corresponding
atomic operation, realized state effect, and targeted failure mode.
}
\label{tab:app_phase_decisions}
\end{table*}

\subsubsection{Historical-Context Recovery}
\label{app:qualitative_history_recovery}

The historical-context recovery scenarios in Phase B make
the distinction between \(M_t\) and \(H_t\) explicit. The initial
\(C_0\) is distractor-rich, while \(H_0\) contains same-task history.
Retrieve cannot recover evidence that was never written to \(M_t\).
SelectEpisode instead identifies the relevant same-task fragment in
\(H_t\) and restores it to \(C_t\), subject to the configured
active-context budget.

Filter may then remove distractors, while Summarize may compress the
restored evidence before the task solver continues. This sequence
recovers earlier task evidence without treating episodic history as
another persistent memory store.

During Phase C, historical-context recovery can occur between LTM
maintenance and final reasoning. Early Add or Update decisions may
preserve persistent knowledge, while a later
SelectEpisode restores an observation or tool output that was
useful only after the task focus changed. The combined state
transition tests whether the policy coordinates \(M_t\),
\(C_t\), and \(H_t\) rather than relying on one memory source.

\subsubsection{Interpretation of Local and Composite-Global Credit}
\label{app:qualitative_credit_analysis}

The two credit branches answer different questions. The local verifier
evaluates whether the current executable command \(u_t\) is relevant,
grounded, useful, and faithful in \(x_t\). The composite global branch
combines task correctness, supporting-fact recall, efficiency, and the
global verifier's evidence-coherence and terminal-memory-consistency
scores. Table~\ref{tab:app_credit_patterns} shows how their normalized
advantages treat common trajectory patterns.

\begin{table*}[t]
\centering
\small
\setlength{\tabcolsep}{5pt}
\renewcommand{\arraystretch}{1.05}
\begin{tabular}{
@{}
p{0.26\textwidth}
p{0.18\textwidth}
p{0.18\textwidth}
p{0.29\textwidth}
@{}}
\toprule
\textbf{Observed Pattern}
&
\textbf{Local Advantage}
&
\textbf{Composite-Global Advantage}
&
\textbf{Credit Interpretation}
\\
\midrule

The command is justified and the completed trajectory is
successful.
&
Positive
&
Positive
&
The command and the completed memory strategy are
reinforced.
\\

The command is justified, but the trajectory later fails.
&
Positive
&
Low or negative
&
The useful transition retains local credit, while the
trajectory-level signal limits its total update.
\\

The command is redundant or weakly grounded, but later
steps still produce a correct answer.
&
Low or negative
&
Positive
&
The command does not inherit the full positive credit of the
successful trajectory.
\\

The proposed command fails a precondition or hard
constraint.
&
Zero
&
Trajectory-dependent
&
The transition is rejected, \(A_k^{\mathrm{global}}\) is retained, and
\(c_{t,k}\) is assigned directly to the corresponding decision.
\\
\bottomrule
\end{tabular}
\caption{
Qualitative interpretation of local and composite-global credit. The
two advantages distinguish command quality from completed-task utility.
}
\label{tab:app_credit_patterns}
\end{table*}

The second and third rows explain the main role of
hierarchical credit assignment. Trajectory-only supervision
cannot separate a useful command followed by a later
failure from a poor command hidden inside a successful
trajectory. Operation-wise normalization further reduces the risk that
tool-specific score ranges dominate the update.

\subsubsection{Failure Modes}
\label{app:failure_analysis}

The failure modes in Table~\ref{tab:app_failure_modes}
follow directly from the seven tool semantics and the
constraint channel. Some errors are semantic and reduce
\(r_{t,k}^{\mathrm{local}}\). Hard failures reject the transition
and contribute to \(c_{t,k}\).

\begin{table*}[t]
\centering
\footnotesize
\setlength{\tabcolsep}{4pt}
\renewcommand{\arraystretch}{1.05}
\begin{tabular}{
@{}
p{0.20\textwidth}
p{0.30\textwidth}
p{0.20\textwidth}
p{0.22\textwidth}
@{}}
\toprule
\textbf{Failure}
&
\textbf{Risk or Observable Consequence}
&
\textbf{Training Response}
&
\textbf{Execution Behavior}
\\
\midrule

Duplicate Add
&
If committed, equivalent information would be stored more than once
and could later introduce redundant context.
&
The rejected command has \(A_{t,k}^{\mathrm{local}}=0\), retains
\(A_k^{\mathrm{global}}\), and receives \(c_{t,k}=1.0\).
&
Reject the transition, keep \(M_t\) and \(C_t\) unchanged, and record
the failure in \(H_t\).
\\

Unsupported Update
&
If committed, the active entry would be revised without visible
evidence and could conflict with persistent memory.
&
The rejected command has \(A_{t,k}^{\mathrm{local}}=0\), retains
\(A_k^{\mathrm{global}}\), and receives \(c_{t,k}=1.0\).
&
Reject the transition, preserve the previous version, and record the
failure in \(H_t\).
\\

Unjustified Delete
&
If committed, useful persistent evidence would become unavailable to
ordinary retrieval.
&
The rejected command has \(A_{t,k}^{\mathrm{local}}=0\), retains
\(A_k^{\mathrm{global}}\), and receives \(c_{t,k}=1.0\).
&
Reject the transition, preserve the active entry, and record the
failure in \(H_t\).
\\

Irrelevant Retrieve
&
Outdated or task-irrelevant LTM content expands \(C_t\) and
raises online cost.
&
Low relevance and local progress; the global efficiency term
may also reduce the composite global score. No hard cost is applied
when the command remains executable.
&
Commit only when the operation remains valid and within
budget.
\\

Over-aggressive Filter
&
If protected evidence were removed, the resulting \(C_t\) could leave
a later reasoning gap.
&
The hard-rejected command has \(A_{t,k}^{\mathrm{local}}=0\), retains
\(A_k^{\mathrm{global}}\), and receives \(c_{t,k}=1.0\).
&
Reject the transition, preserve \(C_t\), and record the failure in
\(H_t\) when protected evidence is selected.
\\

Lossy Summarize
&
If committed, the shorter context would alter a supported relation,
source, or condition needed by the task.
&
The hard-rejected command has \(A_{t,k}^{\mathrm{local}}=0\), retains
\(A_k^{\mathrm{global}}\), and receives the configured hard cost.
&
Reject the transition, preserve \(C_t\), and record the failure in
\(H_t\) when the summary changes the supported meaning.
\\

Misleading SelectEpisode
&
Recent or semantically similar distractors return to \(C_t\)
instead of the relevant earlier fragment.
&
Low relevance and local progress; the completed evidence chain may
also lower the composite global score. No hard cost is applied when
the selected fragments are valid same-task references.
&
Commit only same-task fragments that fit the active-context
budget.
\\

Unnecessary operation
&
The policy adds tool calls and context changes when
\(\varnothing\) would preserve a valid state.
&
Low local progress and potentially lower composite trajectory
efficiency; no hard cost is applied when the command is executable.
&
The transition may remain executable, but receives reduced
credit.
\\
\bottomrule
\end{tabular}
\caption{
Representative failure modes and their treatment by local semantic
verification, composite-global credit, and hard constraints.
}
\label{tab:app_failure_modes}
\end{table*}

The distinction between semantic and hard failures is
important. The irrelevant Retrieve can remain structurally valid and
therefore receive a low semantic score rather than automatic rejection.
Minor but executable omissions in Filter or Summarize are handled the
same way. By contrast, unsupported Update, unjustified Delete,
protected-evidence removal, and meaning-changing Summarize violate an
execution requirement. They are rejected, receive no local semantic
evaluation, retain trajectory-level global credit, and receive a direct
constraint cost. This separation keeps the semantic and hard-constraint
penalties in distinct channels.

\subsubsection{Relation to the Quantitative Results}
\label{app:qualitative_quantitative_link}

The qualitative patterns are consistent with the experiments
in the main paper. Figure~\ref{fig:component_ablation} shows
that LTM operations provide the first major gain, while the
complete LTM/STM command space adds further improvement.
This progression matches the decisions in
Table~\ref{tab:app_phase_decisions}. Persistent storage is
useful only when the policy can also retrieve, filter,
summarize, or restore information at the correct time.

Table~\ref{tab:verifier_ablation} shows complementary gains
from the local and composite-global credit branches. The credit patterns in
Table~\ref{tab:app_credit_patterns} illustrate this distinction. Local
credit distinguishes the quality of one realized transition. The global
verifier evaluates evidence coherence and terminal-memory consistency,
while the composite global branch combines those scores with task
correctness, supporting-fact recall, and efficiency. The observed gains
support complementary roles for operation-level and trajectory-level
credit.

The reward ablation in Table~\ref{tab:reward_ablation}
reports higher \(J_{\mathrm{judge}}\) and MQ, lower TN, and more
memory-tool calls under All-Returns. This pattern is inconsistent with
uniform tool-call suppression as the sole explanation for the lower
token count, but it does not identify operation-specific contributions.
The stronger
efficiency frontier in Figure~\ref{fig:efficiency_frontier}
provides the corresponding system-level result. Detailed
numerical decompositions are reported in
Appendix~\ref{app:additional_quantitative_results}.

\clearpage

\end{document}